\documentclass{article}

\usepackage[preprint]{style}
\usepackage{overpic}
\usepackage[utf8]{inputenc} 
\usepackage[T1]{fontenc}    
\usepackage{hyperref}       
\usepackage{url}            
\usepackage{booktabs}       
\usepackage{amsfonts}       
\usepackage{nicefrac}       
\usepackage{microtype}      
\usepackage{xcolor}         
\usepackage{pifont} 
\usepackage{amsmath}
\usepackage{wrapfig}
\usepackage{placeins} 
\usepackage{caption}
\usepackage{subcaption}

\usepackage{enumitem}
\usepackage{multirow}
\def\ie{\textit{i.e.}}
\def\eg{\textit{e.g.}}

\newcommand{\figref}[1]{Fig.~\ref{#1}}

\newcommand{\secref}[1]{Sec.~\ref{#1}}

\newcommand{\name}{{\textit{\textbf{HOComp}}}}

\title{HOComp: Interaction-Aware Human-Object Composition}

\author{%
  Dong Liang \\
  Tongji University / CityUHK \\
  \texttt{sse\_liangdong@tongji.edu.cn} \\
  \And
  Jinyuan Jia\thanks{Joint corresponding authors.} \\
  Tongji University / HKUST(GZ) \\
  \texttt{jinyuanjia@hkust-gz.edu.cn} \\
  \And
  Yuhao Liu\footnotemark[1] \\
  CityUHK \\
  \texttt{yuhaoliu7456@gmail.com} \\
  \And
  Rynson W.H. Lau\footnotemark[1] \\
  CityUHK \\
  \texttt{Rynson.Lau@cityu.edu.hk} \\
}

\begin{document}

\maketitle

\begin{abstract}
While existing image‑guided composition methods may help insert a foreground object onto a user-specified region of a background image, achieving natural blending inside the region with the rest of the image unchanged, we observe that these existing methods often struggle in synthesizing seamless interaction-aware compositions when the task involves human-object interactions.
In this paper, we first propose \name, a novel approach for compositing a foreground object onto a human-centric background image, while ensuring harmonious interactions between the foreground object and the background person and their consistent appearances. 
Our approach includes two key designs: (1) \textit{\textbf{M}LLMs-driven \textbf{R}egion-based \textbf{P}ose \textbf{G}uidance (MRPG)}, which utilizes MLLMs to identify the interaction region as well as the interaction type (\eg, holding and lefting) to provide coarse-to-fine constraints to the generated pose for the interaction while incorporating human pose landmarks to track action variations and enforcing fine-grained pose constraints; and (2) \textit{\textbf{D}etail-\textbf{C}onsistent \textbf{A}ppearance \textbf{P}reservation (DCAP)}, which unifies a shape-aware attention modulation mechanism, a multi-view appearance loss, and a background consistency loss to ensure consistent shapes/textures of the foreground and faithful reproduction of the background human. 
We then propose the first dataset, named \textit{Interaction-aware Human-Object Composition (IHOC)}, for the task. Experimental results on our dataset show that \name ~effectively generates harmonious human-object interactions with consistent appearances, and outperforms relevant methods qualitatively and quantitatively. 
Project page:  \url{https://dliang293.github.io/HOComp-project/}.

\end{abstract}

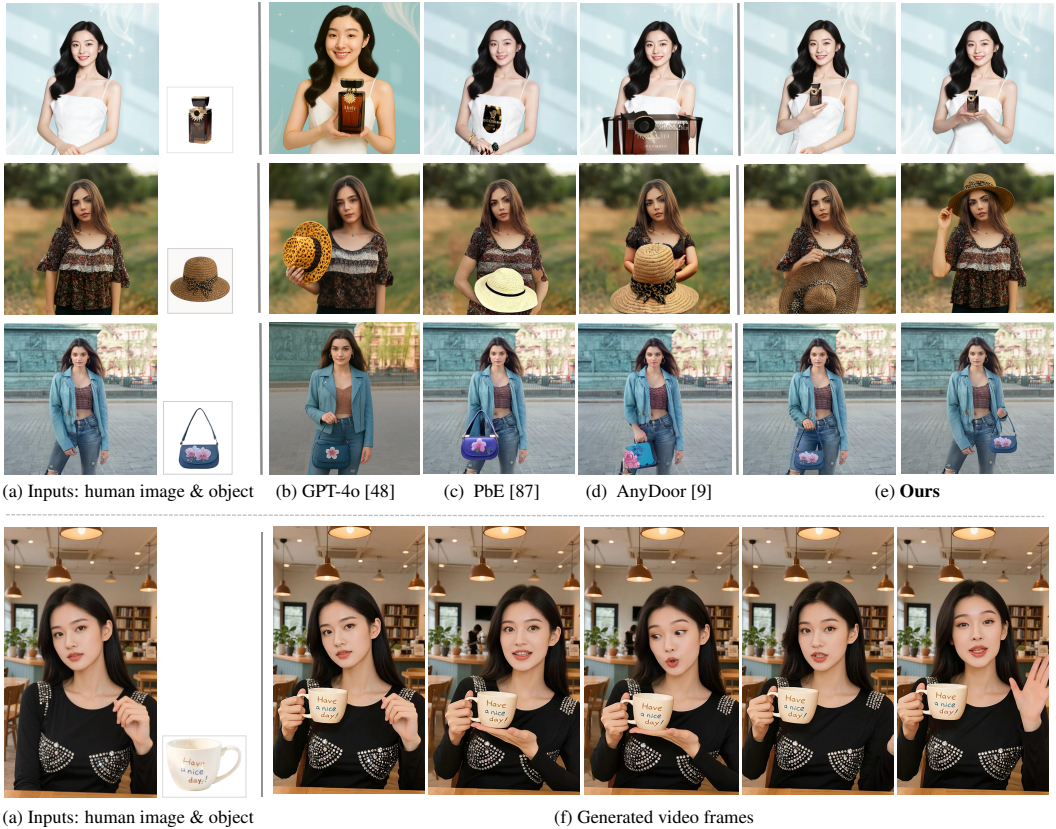
\begin{figure}[htbp]
    \centering
\begin{overpic}[width=1\textwidth]{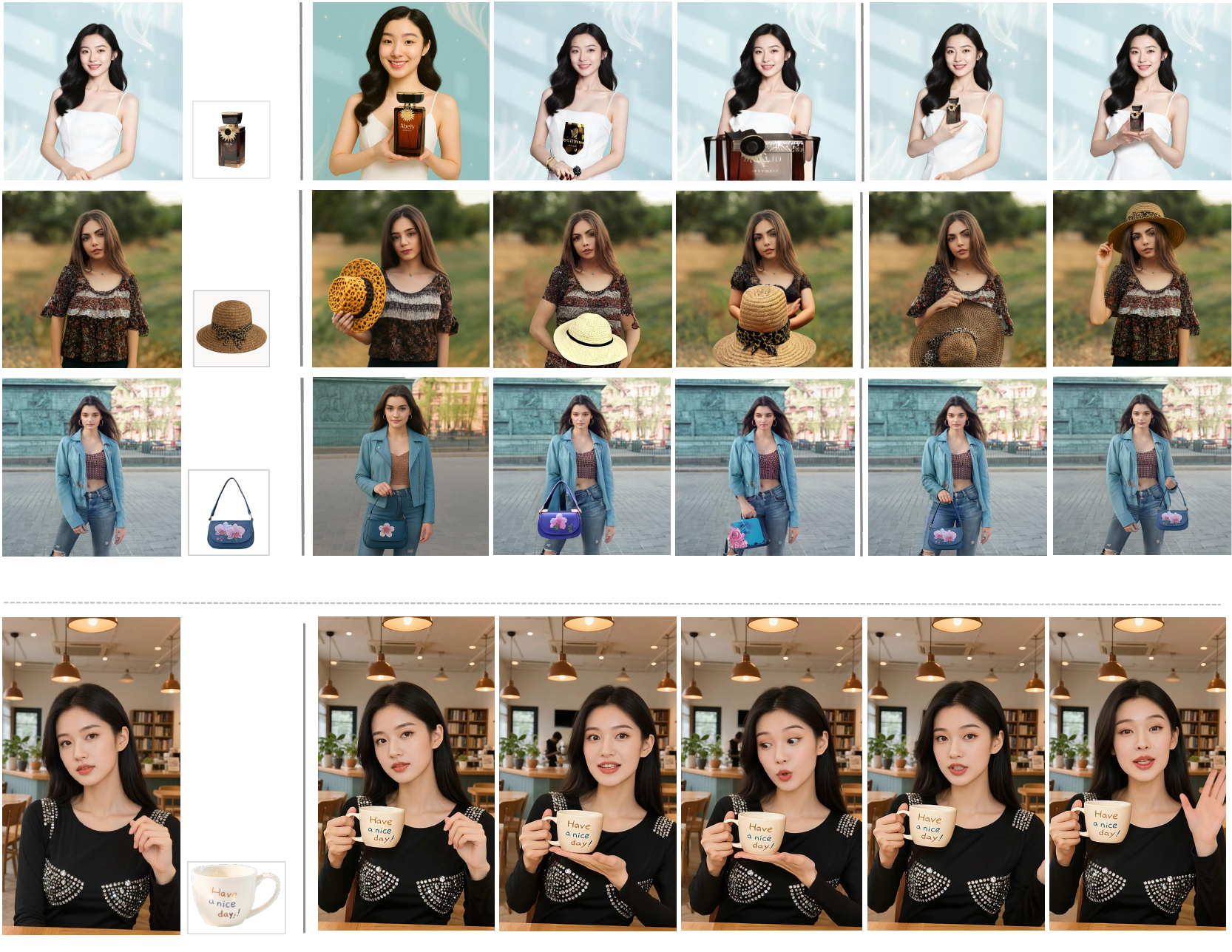}
    \put(0,30){\scriptsize{(a) Inputs: human image \& object}}
      \put(26,30){\scriptsize{(b) GPT-4o~\cite{chatgpt}}}
      \put(42,30){\scriptsize{(c) ~PbE~\cite{yang2023paint}}}
      \put(55.5,30){\scriptsize{(d) ~AnyDoor~\cite{chen2024anydoor}}}
      \put(83,30){\scriptsize{(e) \textbf{Ours}}}
      \put(0,-1){\scriptsize{(a) Inputs: human image \& object}}
      \put(52.5,-1){\scriptsize{(f) Generated video frames}}
    \end{overpic}
    \vspace{1mm}
    \caption{When compositing a foreground object onto a human-centric background image, existing methods (b-d) typically rely on manually specifying the target region and text prompt, and often produce unrealistic interactions and inconsistent foreground/background appearances. In contrast, our proposed ~\name, automatically identifies the target region and generates a suitable text prompt to guide the interaction, resulting in realistic, harmonious and diverse interactions. Note that the text prompts used by the existing methods in the above three examples are: ``A model is showing a perfume bottle'', ``A girl is holding a hat'', and ``A woman is lifting a handbag''. By integrating with an Image-to-Video (I2V) model, our approach can support applications like human-product demonstration video generation (see results on the bottom region).}
\label{fig:teaser}
\vspace{-3mm}
\end{figure}

\section{Introduction}
\label{sec:intro}

Considering a scenario in which a designer aims to create a perfume advertisement by compositing the image of a product onto an existing photograph with a human person, as shown in row 1 of Fig.~\ref{fig:teaser}, two critical objectives need to be satisfied in order to produce a visually convincing output. 
First, the interaction between the person and the perfume bottle should appear \textbf{\textit{natural}}, such that the bottle may seem to be appropriately related to (\eg, held by) the person. 
Second, visual \textbf{\textit{consistency}} must be maintained, preserving the original identities of both the person (including facial features and makeup) and the perfume bottle (\eg, the logo, color, and shape).

Some existing image-guided composition tasks~\cite{xue2022dccf,li2024tuning,winter2024objectdrop} may be most relevant to the above task setting. They take a user-supplied foreground exemplar, typically accompanied by a textual prompt and a user-defined target region, and aim to synthesize a harmonious composition.
Within this paradigm, they either incorporate identity-preservation modules~\cite{chen2024anydoor,song2024imprint} to explicitly retain the original foreground details or focus on adjusting the colors, shadows, and perspective of the foreground to harmonize it with the background~\cite{lu2023tf, tarres2024thinking, yang2023paint, song2023objectstitch}, thereby producing photorealistic compositions. 
Despite the success, when the composition involves human and object interactions, as depicted in Fig.~\ref{fig:teaser}, existing methods~\cite{chen2024anydoor,song2023objectstitch,yang2023paint} struggle to produce satisfactory results.

For our composition task, we observe that existing methods tend to fail in one or both of the following ways: (1) they may produce inappropriate gestures for the background persons (\eg, most results in Fig.~\ref{fig:teaser}(c,d)); and (2) they may change the contents/identities of the foreground objects (\eg, rows 2 and 3 of Fig.~\ref{fig:teaser}(b-d)) and/or the background persons (\eg, the face in row 1 of Fig.~\ref{fig:teaser}(b), and the clothes in row 2 of Fig.~\ref{fig:teaser}(b,c) and row 3 of Fig.~\ref{fig:teaser}(b,d).
To address these problems, we propose \name, an interaction-aware human-object composition framework, to create seamless composited images with harmonious human-object interactions and consistent appearances.

Our \name ~includes two key designs. The first design is the \textit{MLLMs-driven region-based pose guidance (MRPG)}, which aims to constrain the human-object interaction. 
By utilizing the capabilities of MLLMs, our method automatically determines suitable interaction types~\footnote{This interaction type is embedded in the text prompt. 
For example, ``A woman is \textit{\textbf{holding}} a hat'', and ``A kid is \textit{\textbf{eating}} a donut.''} (\eg, \textit{holding, eating}) and interaction region. Here, we adopt a \textit{coarse-to-fine constraint strategy}. 
We first use the interaction region generated by MLLMs as a coarse-level constraint to restrict the region of the background image for the interaction. 
We then incorporate human pose landmarks as a supervision to capture the variation of the human pose in the interaction, providing a fine-grained constraint on the pose within the interaction region. 
The second design is the \textit{detail-consistent appearance preservation (DCAP)}, which aims to ensure foreground/background appearance consistency.
For the foreground object, we propose a shape-aware attention modulation mechanism to explicitly manipulate attention maps for maintaining a consistent object shape, and a multi-view appearance loss to further preserve the object textures at the semantic level. 
For the background image, we propose a background consistency loss to retain the details of the background person outside the interaction region.

To train the model, we introduce a new dataset called \textit{Interaction-aware Human-Object Composition (IHOC) dataset}, which includes images of humans before and after interacting with the foreground object, the interaction region, and the corresponding interaction type. 
We conduct extensive experiments on this dataset, and the results demonstrate that our approach can generate accurate and harmonious human-object interactions, resulting in highly realistic and convincing compositions.

The main contributions of this work include:
\begin{enumerate}
    \item We propose a new approach for interaction-aware human-object composition, named \textbf{\textit{HOComp}}, which focuses on seamlessly integrating a foreground object onto a human-centric background image while ensuring harmonious interactions and preserving the visual consistency of both the foreground object and the background person.

    \item \name ~incorporates two innovative designs: \textit{MLLMs-driven region-based pose guidance (MRPG)} for constraining human-object interaction via a \textit{coarse-to-fine} strategy, and \textit{detail-consistent appearance preservation (DCAP)} for maintaining consistent foreground/background appearances.
    
    \item We introduce the \textit{Interaction-aware Human-Object Composition (IHOC) dataset}, and conduct extensive experiments on this dataset to demonstrate the superiority of our method.

\end{enumerate}

\section{Related Works}
\label{sec:related_works}

\textbf{Image-guided Composition.} 
It aims to seamlessly integrate a user-provided foreground exemplar onto a designated region of a background image, sometimes with textual guidance. Existing methods either focus on appearance harmonization (\ie, adjusting colors, shadows, and perspective) in order to integrate the foreground onto the background seamlessly \cite{pham2024tale,yu2025omnipaint, chen2024zero, ren2024relightful,tao2024motioncom, ruiz2024magic, liang2025vodiff,wang2024primecomposer,he2024affordance,chen2024mureobjectstitch} or emphasize identity preservation by introducing dedicated modules to maintain the identity consistency of the object across scenes \cite{chen2024anydoor,song2024imprint,winter2024objectmate,li2024textit,zhang2023controlcom,song2025insert}. 
However, these methods primarily refine the foreground and often fail to generate natural, realistic human gestures or poses in human-object interactions (HOIs). While DreamFuse~\cite{huang2025dreamfuse} adjusts the foreground to adapt to the background context, it supports only limited hand actions and struggles with complex HOIs. 
With the advance of DiT models~\cite{peebles2023scalable}, recent works~\cite{tian2025mige, wang2025unicombine, xiao2024omnigen,wang2024genartist,batifol2025flux} propose unified frameworks to integrate multiple image generation/editing tasks. Similar to multi-modality methods~\cite{xai2025grok3, chatgpt, liu2025step1x}, these approaches often unintentionally modify the background human and introduce inconsistencies in the foreground object.

\noindent \textbf{Multi-Concept Customization.} It aims to generate images that align with both the text prompt and user-specified concepts, facilitating the creation of personalized content. 
Tuning-based methods~\cite{kumari2023multi,avrahami2023break,tewel2023key,liu2023customizable,liu2023cones,gu2023mix,lin2024non} typically incorporate new concepts into diffusion models by fine-tuning specific parameters, but each new concept requires a separate tuning process. Instead, 
training-based methods~\cite{xiao2024fastcomposer, patellambda,zhang2024ssr,wangms,kong2024omg,chen2024unireal,lin2025multitwine,deng2025cinema,liang2025movie} train additional modules to extract the identity of a concept and inject it into the denoising network via attention layers. 
Training-free methods~\cite{ding2024freecustom, wang2024magicface, yao2025freegraftor, woo2025flipconcept} incorporate reference-aware attention mechanisms. These methods typically re-generate both the foreground object and background human, leading to inconsistent background human appearance.

\noindent \textbf{Human-Object Interaction (HOI) Generation.} 
It aims to synthesize images that depict plausible and coherent interactions between humans and objects. 
Recent diffusion-based methods generate HOI images by introducing extra cues, such as bounding boxes~\cite{hoe2024interactdiffusion,jiang2024record, hua2021exploiting} or pose structures~\cite{zhang2023adding,li2023gligen,chen2024virtualmodel}, reference videos~\cite{xu2024anchorcrafter,wei2025dreamrelation}, and in-context samples of similar interactions~\cite{huang2024learning,zhang2023motioncrafter,huang2024reversion}. 
However, all these approaches require additional inputs during inference (\eg, human poses or images describing the interaction). Some works~\cite{ye2023affordance,xue2024hoi} adjust human hand poses during interactions, but this is often insufficient for complex scenarios. 
Other methods~\cite{shi2025dreamrelation, hu2025personahoi, parihar2024text2place, gafni2020wish, yang2024person} employ relation-aware frameworks to improve HOI generation in subject-driven settings, yet they fail to preserve the background human appearance consistency.
Concurrent works, DreamActor-H1~\cite{wang2025dreamactor} and HunyuanVideo-HOMA~\cite{huang2025hunyuanvideo}, explore human interaction in the contexts of human-product demonstrations and animated human-object interactions. They incorporate additional modality guidance and exploit the strong multi-modal fusion capabilities of the DiT framework for video generation.

In summary, existing methods fall short in addressing the challenge of our interaction-aware human-object composition task, which requires the model to produce harmonious human-object interactions and consistent foreground/background appearances.

\section{Method}

\begin{figure}[t]
    \centering
    \begin{overpic}[width=1\textwidth]{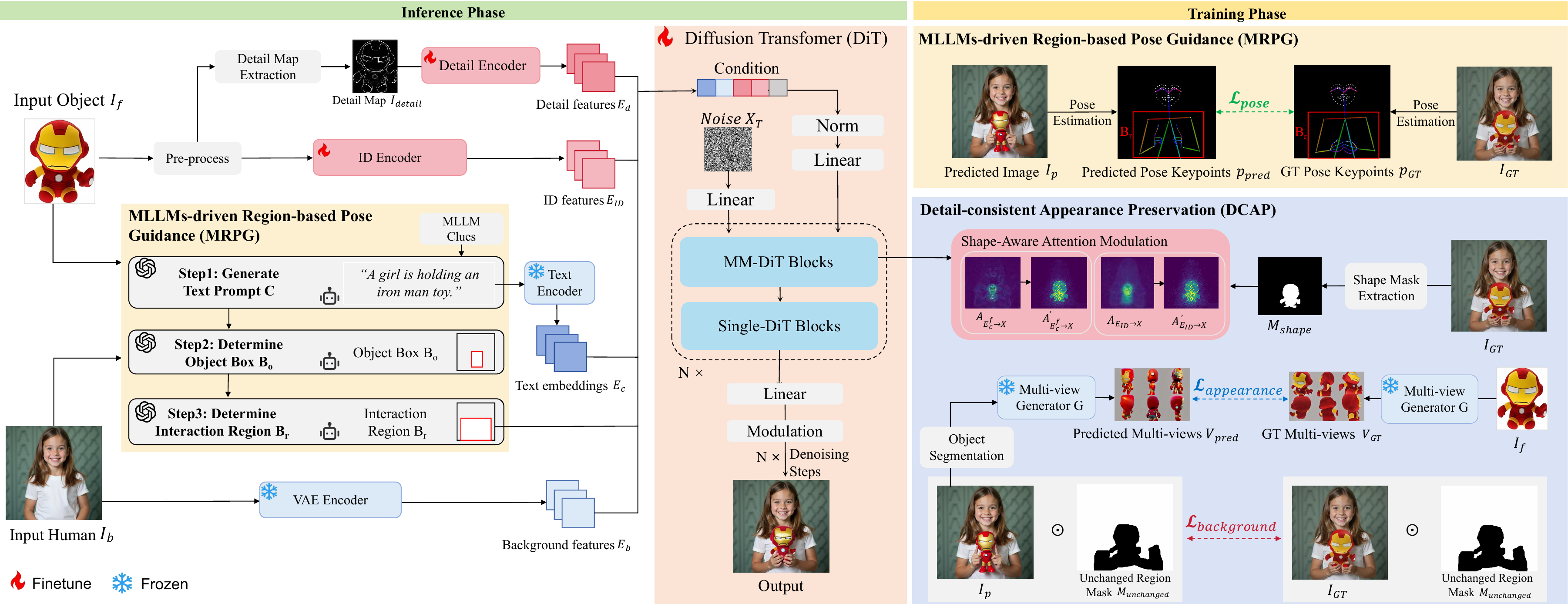}
    \end{overpic}
        \caption{
        \textbf{Pipeline of \name.} Our method includes two core modules: MRPG for constraining human-object interaction and DCAP for maintaining appearance consistency. 
        \textbf{Inference Phase (left):} MRPG uses MLLMs to generate a text prompt $C$, object box $B_o$ and interaction region $B_r$.  Among these, \(B_r\) and \(C\) are encoded and, together with the object ID, detail features, and background features, are used to condition the DiT for final composition generation.
        \textbf{Training Phase (right):} MRPG constrains the interaction by applying a \textit{pose-guided loss} $\mathcal{L}_{\text{pose}}$ with keypoint supervision. DCAP enforces appearance consistency via: (1) \textit{shape-aware attention modulation} to adjust the attention maps to follow the object’s shape prior $M_\text{shape}$; (2) a multi-view appearance loss $\mathcal{L}_{\text{appearance}}$ to semantically align synthesized and input foregrounds (multi-views); and (3) a background loss $\mathcal{L}_{\text{background}}$ to preserve original background details.
    }
    \label{fig:pipe}
\vspace{-5mm}
\end{figure}

Given a foreground object image $\mathbf{I_f}$ and a background image $\mathbf{I_b}$ containing a human subject, our goal is to synthesize a harmoniously composited image $\mathbf{I_p}$ that integrates the foreground object onto the human-centric background image. The composited image should exhibit harmonious interactions and maintain appearance consistency between the foreground object and the background human.

To achieve this objective, we propose \name, an interaction-aware human-object composition framework, as illustrated in Fig.~\ref{fig:pipe}. Our framework includes two key components: \textit{MLLM-driven Region-based Pose Guidance (MRPG)} and \textit{Detail-Consistent Appearance Preservation (DCAP)}. MRPG leverages Multimodal Large Language Models (MLLMs) and human pose priors to constrain human-object interaction in a coarse-to-fine manner. DCAP preserves the shape and texture of the foreground object while maintaining details of the background human, ensuring faithful and coherent appearance reproduction throughout the composited scene.

In the remainder of this section, we first introduce the preliminaries in Sec.~\ref{subsec:pre}. We then detail the design of MRPG in Sec.~\ref{subsec:pose_guidance}, followed by DCAP in Sec.~\ref{subsec:fidelity}. Finally, we describe our Interaction-aware Human-Object Composition (IHOC) dataset in Sec.~\ref{subsec:dataset}.

\subsection{Preliminary}
\label{subsec:pre}

\noindent \textbf{Diffusion Transformer} (DiT) is a transformer-based diffusion model for image synthesis. Given a noisy latent $\mathbf{z}_t$ at timestep $t$, it predicts the denoised output via $\hat{\mathbf{z}}_0 = \text{DiT}(\mathbf{z}_t, t, c)$, where $c$ denotes a conditioning signal (\textit{e.g.}, text embeddings or visual prompts). Owing to its scalability and strong generative capacity, DiT serves as a robust backbone for conditional image generation.

\noindent \textbf{Attention Manipulation} is a key strategy for improving semantic alignment and structural control in diffusion models through attention map editing, external signal injection, or modified attention weight computation. For a standard attention layer defined as $\mathbf{A} = \mathrm{softmax}(\mathbf{Q}\mathbf{K}^\top / \sqrt{d}) \mathbf{V}$, manipulation introduces a structured bias or conditioning modulation:
$\mathbf{A}' = \mathrm{softmax}((\mathbf{Q}\mathbf{K}^\top + \mathbf{M}) / \sqrt{d}) \mathbf{V}$, where $\mathbf{M} \in \mathbb{R}^{n \times n}$ encodes spatial priors or prompt-specific relevance (e.g., object masks).

\subsection{MLLM-driven Region-based Pose Guidance (MRPG)}
\label{subsec:pose_guidance}

MRPG adopts a coarse-to-fine strategy to constrain the human-object interaction. At the coarse level, it leverages the reasoning capabilities of MLLMs to automatically identify suitable interaction type and corresponding interaction region through a multi-stage querying process. At the fine level, a \textit{pose-guided loss} is introduced to impose fine-grained constraints on human poses within the interaction region, explicitly supervising the predicted image using human pose keypoints.

\noindent \textbf{Generating Interaction Regions and Types.} As illustrated in Fig.~\ref{fig:pipe}, we employ MLLMs (\textit{e.g.,} GPT-4o) in a chain-of-thought, a step-by-step process to generate the interaction type (denoted as a text prompt \(C\)) and the interaction region (represented by a bounding-box \(B_r\)). While the interaction type specifies what interaction is to be performed by the background person on the foreground object (\eg, holding), the interaction region specifies the location in the image that the interaction is to be performed. Specifically, we send the foreground object and the background image to the MLLM and query it in a three-stage approach:
(1) With a set of initial prompts as the instruction guidance, we ask the MLLM to envision a plausible interaction type and return the interaction type in the form of a text prompt description \(C\);
(2) Conditioned on \(C\), we ask the MLLM to further infer a potential region (\ie, bounding box \(B_o\)) in the background image where the foreground object is to be placed; 
(3) We ask the MLLM to identify the interaction region \(B_r\) by considering which human body parts are involved in the interaction. The generated interaction region \(B_r\) is converted into a mask, encoded via a VAE~\cite{kingma2013auto}, and used alongside text embeddings \(E_c\) as conditioning inputs to the DiT model. 

\noindent \textbf{Imposing Fine-grained Pose Guidance.} Considering the significant correlation between human-object interactions and body poses, we introduce a pose-guided loss \(\mathcal{L}_{pose}\) to impose fine-grained constraints on poses within the interaction region. Let \(\mathbf{p}^i_{\text{GT}}\) and \(\mathbf{p}^i_{\text{pred}}\) represent the \(i\)-th keypoint detected by a pose estimator $\mathbf{G}_p$ from the ground-truth image \(I_{\text{GT}}\) and the predicted image \(I_p\), respectively. The pose-guided loss \(\mathcal{L}_p\) is formulated as:
\begin{equation}
\mathcal{L}_{p} = \frac{1}{n} \sum_{i \in B_r} \left\| \mathbf{p}^i_{\text{GT}} - \mathbf{p}^i_{\text{pred}} \right\|^2,
\end{equation}
where \(n\) denotes the number of pose keypoints located within the interaction region \(B_r\), as illustrated in Fig.~\ref{fig:pipe}. This localized pose-guided loss explicitly directs the model’s optimization efforts towards accurately capturing human poses involved in the interaction, rather than globally adjusting the entire human pose, thereby enhancing the realism and harmony of the generated interaction.

\subsection{Detail-Consistent Appearance Preservation (DCAP)}
\label{subsec:fidelity}

To ensure fine-grained appearance consistency, for the \textbf{foreground}, we first extract identity and detail information as conditioning inputs for the DiT model. To enforce shape consistency, we introduce a \textit{shape-aware attention modulation} mechanism to adjust the foreground-relevant attention maps in the MM-DiT blocks, guiding the attention maps to align with the foreground object’s shape prior better. For texture consistency, we propose a \textit{multi-view appearance loss} to maintain semantic alignment across multiple viewpoints. For the \textbf{background}, we leverage an unchanged region mask to identify unaffected areas and impose a \textit{background consistency loss} to preserve original background details.

\noindent \textbf{Foreground Object Identity and Detail Extraction.}
We first preprocess the foreground object by removing the background and centering it. To capture the identity information, we then employ the DINOv2-based ID encoder~\cite{oquabdinov2}, renowned for robust semantic representations, to extract the foreground ID features $E_{ID}$. As the resulting identity tokens have a coarse spatial resolution and therefore lack texture details, we extract a high-frequency detail map, $I_{\text{detail}}$, as an additional condition: $I_{\text{detail}} = I_{\text{gray}} - \operatorname{GaussianBlur}(I_{\text{gray}})$, where \(I_{\text{gray}}\) is the grayscale foreground image. A lightweight detail encoder~\cite{chen2024anydoor} processes $I_{\text{detail}}$ to extract detail features $E_{d}$, which are then fused with foreground ID features $E_{ID}$ to condition the DiT model.

\begin{wrapfigure}{r}{0.37\textwidth}
    \centering
    \vspace{-28pt}
    \begin{overpic}[width=0.37\textwidth]{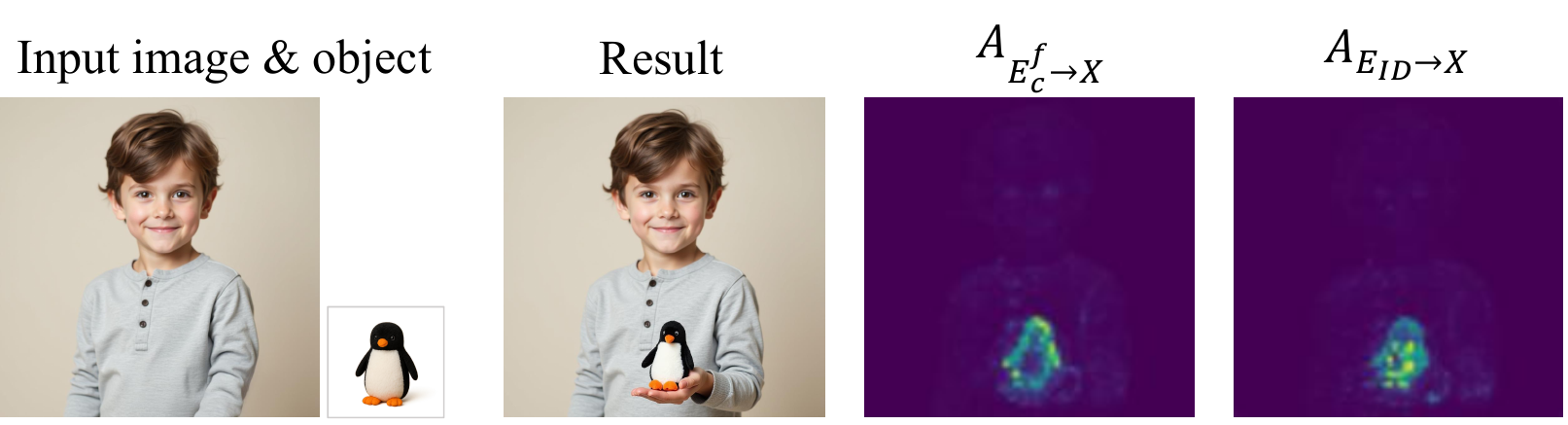}
    \end{overpic}
    \caption{Visualization of attention maps related to the foreground text embeddings \( A_{\mathbf{E^f_{c}} \rightarrow \mathbf{X}} \) and the identity features \( A_{\mathbf{E}_{\text{ID}} \rightarrow \mathbf{X}} \) , both exhibiting strong alignment with object shape.}
    \label{fig:attention}
    \vspace{-20pt}
\end{wrapfigure}

\noindent \textbf{Shape-aware Attention Modulation.}
To enhance shape consistency, we modulate foreground-relevant attention maps in the MM-DiT blocks, encouraging the attention maps to align more precisely with the object’s shape prior. This design is motivated by the observation that these attention maps highlight object shapes (see Fig.~\ref{fig:attention}), indicating that the model is able to capture structural cues of the foreground objects.

Specifically, we compute two foreground-relevant attention maps: one based on the foreground ID features \( \mathbf{E_{ID}} \), and the other on the foreground text embeddings \( \mathbf{E^f_{c}} \), with \( \mathbf{X} \) denoting the target image tokens. Here, \( \mathbf{E^f_{c}} \) are extracted from the full text embedding \( \mathbf{E_c} \). For instance, if \emph{"toy"} is annotated as a foreground object in the text prompt $C$ \emph{"A boy is holding a toy"}, \( \mathbf{E^f_{c}} \) is the sub-embedding aligned with \emph{"toy"} from \( \mathbf{E}_c \). The attention maps are computed as:

\begin{equation} 
A_{\mathbf{E^f_{c}} \rightarrow \mathbf{X}} = \operatorname{softmax} \left( \frac{Q_{\mathbf{X}} K_{\mathbf{E^f_{c}}}^\top}{\sqrt{d}} \right), \quad A_{\mathbf{E_{ID}} \rightarrow \mathbf{X}} = \operatorname{softmax} \left( \frac{Q_{\mathbf{X}} K_{\mathbf{E_{ID}}}^\top}{\sqrt{d}} \right), 
\end{equation} 
where \( Q_{\mathbf{X}} \in \mathbb{R}^{N \times d} \) are queries from the target image tokens, and \( K_{\mathbf{E^f_{c}}}, K_{\mathbf{E_{ID}}} \in \mathbb{R}^{M \times d} \) are keys projected from \( \mathbf{E^f_{c}} \) and \( \mathbf{E_{ID}} \), respectively.

To obtain the shape prior, as shown in Fig.~\ref{fig:pipe}, we extract a foreground object mask \( M_{\text{shape}} \) from the ground-truth image. We aim to enhance the attention within the object region while suppressing distractions outside it. 
Considering that directly modifying the attention maps may potentially compromise the image quality of the pre-trained model~\cite{kim2023dense}, we adopt a residual-based modulation strategy over the extracted attention maps \( A_{\mathbf{E^f_{c}} \rightarrow \mathbf{X}} \) and \( A_{\mathbf{E_{ID}} \rightarrow \mathbf{X}} \) to incorporate shape priors while preserving the original attention distribution. The modulation is defined as:
\begin{equation}
A' = A + \alpha \cdot \left( M_{\text{shape}} \cdot (A_{\max} - A) - (1 - M_{\text{shape}}) \cdot (A - A_{\min}) \right),
\end{equation}
where \( A \in \{ A_{\mathbf{E^f_{c}} \rightarrow \mathbf{X}}, A_{\mathbf{E_{ID}} \rightarrow \mathbf{X}} \} \). \( A_{\max} \) and \( A_{\min} \) are the per-query maximum and minimum values computed row-wise. The scalar \( \alpha \in \mathbb{R}^{+} \) controls the modulation strength. The modulated attention map is then integrated into the DiT model to encourage shape-aware feature learning.

\noindent \textbf{Multi-view Appearance Loss.}
To address texture inconsistencies caused by changes in viewpoint during interactions, we encourage the predicted foreground object to maintain consistent semantic appearance with the ground truth across diverse views. Specifically, we synthesis multi-view images for both the predicted result and the input foreground, and measure their semantic similarity.

As shown in Fig.~\ref{fig:pipe}, we first segment the predicted foreground object from $\mathbf{I}_p$. Given the segmented output and the input foreground image $\mathbf{I}_f$, we apply a multi-view generator \( G \) to synthesize \( k \) views:
\begin{equation}
\mathbf{V}_{\text{pred}} = \{ \mathbf{V}_{\text{pred}}^{(i)} \}_{i=1}^{k} = G(\operatorname{Segment}(\mathbf{I}_p)), \quad 
\mathbf{V}_{\text{GT}} = \{ \mathbf{V}_{\text{GT}}^{(i)} \}_{i=1}^{k} = G(\mathbf{I}_f).
\end{equation}
We then extract CLIP~\cite{radford2021learning} features from each synthesized view: $\mathcal{F}_{\text{pred}}^{(i)} = \operatorname{CLIP}(\mathbf{V}_{\text{pred}}^{(i)}),  
\mathcal{F}_{\text{GT}}^{(i)} = \operatorname{CLIP}(\mathbf{V}_{\text{GT}}^{(i)}).$
The multi-view appearance loss is then formulated as:
\begin{equation}
    \mathcal{L}_{appearance} = \frac{1}{k} \sum_{i=1}^{k} \left( 1 - \frac{ \mathcal{F}_{\text{pred}}^{(i)} \cdot \mathcal{F}_{\text{GT}}^{(i)} }{ \left\| \mathcal{F}_{\text{pred}}^{(i)} \right\| \left\| \mathcal{F}_{\text{GT}}^{(i)} \right\| } \right),
\end{equation}
which encourages semantic alignment of the predicted object with the ground truth across multi-views.

\noindent \textbf{Background Consistency Loss.} To preserve the appearance of the background human during the process, we utilize an unchanged region mask \( M_{\text{unchanged}} \), which is provided by our dataset and indicates the region that remains unaffected throughout the interaction. By constraining the generated image to match the ground-truth image in this unchanged region, we enforce consistency with the original background appearance. The background consistency loss \( \mathcal{L}_{b} \) is defined as:
\begin{equation}
    \mathcal{L}_{background} = \sum_{i \in I} M_{unchanged}^i \odot \left\| \mathbf{x}_{GT}^i - \mathbf{x}_{pred}^i \right\|^2 ,
\end{equation}
where $\mathbf{x}_{GT}$ and $\mathbf{x}_{pred}$ denote the pixel values of the ground-truth image $\mathbf{I}_{GT}$ and of the predicted image $\mathbf{I}_{p}$, respectively.

\noindent \textbf{Overall Training Objective}.
The model is optimized with the composite loss:
\begin{equation}
\mathcal{L}_{\text{total}}
   = \mathcal{L}_{\text{denoising}}
   + \alpha_1 \mathcal{L}_{p}
   + \alpha_2 \mathcal{L}_{b}
   + \alpha_3 \mathcal{L}_{a},
\end{equation}
where \(\mathcal{L}_{\text{denoising}}\) is the standard denoising loss.
\(\mathcal{L}_{p}, \mathcal{L}_{b}, \mathcal{L}_{a}\) are the pose-guided, background consistency, and multi-view appearance losses.  \(\alpha_1, \alpha_2, \alpha_3\) are the coefficients of the corresponding loss terms.

\subsection{Dataset Preparation} 
\label{subsec:dataset}
We introduce the \textit{Interaction-aware Human-Object Composition (IHOC)} dataset to address the lack of paired pre- and post-interaction data crucial for modeling realistic and coherent human-object compositions. IHOC includes six components: (1) \emph{background human images} (without the object); (2) \emph{foreground object images}; (3) \emph{composited images} with harmonious interactions and consistent appearances; (4) \emph{text prompts} describing the interaction type; (5) \emph{interaction regions}; and (6) \emph{unchanged region masks} to indicate unaffected background areas.

Our dataset is constructed through the following stages: \ding{182}~\textbf{Composited Images:} To enhance data diversity, we adopt the 117 human-object interaction types defined by HICO-DET~\cite{chao2018learning} and include both real and synthetic samples. 
For real data, we manually select 50 images per type (5,850 total) from HICO-DET. To ensure the quality of our dataset and to reduce bias, we exclude images that (1) contain multiple persons, (2) lack clearly visible persons (\eg, only a hand is shown), or (3) have large parts of the foreground objects occluded or not visible (\eg, only one wheel of a bicycle is visible), making it difficult to identify them. The final selection emphasizes diversity in object type, scale, and human pose across diverse scenes. For synthetic data, we use GPT-4o to generate 50 prompts per type and synthesize 5,850 images using FLUX.1 [dev]~\cite{flux}. These synthetic samples help complement the real data by introducing a wider range of human appearances, poses, viewpoints, and visual styles (\eg, cartoon, sketches). In total, we have collected 11,700 composited interaction examples. \ding{183}\textbf{Foreground Object Images:} Foreground objects are segmented from the composite images using SAM~\cite{ravi2024sam}. To address occlusions caused by human-object interactions, we use GPT-4o to infer and complete missing regions, producing plausible and visually consistent object appearances. \ding{184}~\textbf{Background Human Images \& Unchanged Region Masks:} We manually inpaint composite images using FLUX.1 Fill [dev]~\cite{flux-fill} to remove interacting objects and recover plausible human poses without the interactions. An inpainting mask denotes an interaction-altered region; its inverse produces the unchanged region mask, highlighting the area unaffected by the interaction. \ding{185}~\textbf{Text Prompts \& Interaction Regions}: For real images, we use GPT-4o to generate text prompts. For synthetic images, we reuse the generation prompts. In addition, we use GPT-4o to annotate each prompt with foreground object tokens, indicating which words correspond to the foreground objects. The interaction regions are derived by inverting the unchanged region masks.  More information on our dataset, including statistics and visualizations, can be found in Sec.~\ref{sec:dataset} of the Appendix.

\section{Experiments}

\noindent \textbf{Implementation Details.}
We adopt FLUX.1 [dev]~\cite{flux} as the base model and fine-tune it using LoRA~\cite{hu2022lora} with rank 16, applied to the attention layers. All training images are resized to 512$\times$512 resolution. The model is trained for 10,000 steps with a batch size of 2, using AdamW and a learning rate of 1e\text{-}5. Training takes approximately 20 hours on 2$\times$A100 GPUs. We employ DWPose~\cite{yang2023effective} for pose estimation, Zero123+~\cite{shi2023zero123++} for multi-view generation and GPT-4o\cite{chatgpt} as MLLM in MRPG.

\noindent \textbf{Evaluation Metrics}.
We use \textbf{FID}~\cite{heusel2017gans} to assess the overall quality of the generated images, where a lower score indicates a better alignment with real images. To evaluate how well a generated image depicts the specified human-object interaction (\ie, HOI Alignment), we compute the \textbf{HOI-Score} using a pre-trained HOI detector (\eg, UPT \cite{zhang2022efficient}), which measures the accuracy of the interaction in the generated image. Additionally, we employ the \textbf{CLIP-Score} \cite{hessel2021clipscore} to evaluate the global semantic alignment between the generated image and the text prompt. 
Subsequently, we use the \textbf{DINO-Score} to assess how well the foreground object appearance is preserved, where a higher score indicates a better appearance consistency to the input foreground object. Finally, background consistency is evaluated by computing the Structural Similarity Index \textbf{(SSIM)} \cite{wang2004ssim} over the area outside the interaction region, where a higher SSIM(BG) score indicates a better retention of the original background.

\noindent \textbf{Benchmark.} 
We introduce a new benchmark, \textbf{HOIBench}, to evaluate the quality of the human-object interaction task. 
We begin by collecting 30 images, each with a human person, from the internet. The humans in these images cover diverse appearances, including different poses and clothes. Half of these images feature the upper body, while the other half depict the full body. To ensure a broad range of interaction types, we adopt the 117 interaction types defined in the HICO-DET~\cite{hou2020visual}.
We prompt GPT-4o with each type to infer a plausible foreground object (\eg, \textit{playing}~$\rightarrow$~\textit{guitar}). A concise textual description of each object is then used to retrieve a representative image from the internet, yielding 117 interaction–foreground image pairs.
Finally, for each human image, we randomly sample 20 interaction-object pairs from the generated set, producing a total of 600 human-object interaction instances (20 interactions × 30 human images) for evaluation.

\begin{table*}[t]
\centering
\small
\caption{Quantitative comparison of our method with nine SOTA methods.  
The user study reports the averaged rank (lower is better) of nine methods in image quality (IQ), interaction harmonization (IH), and appearance preservation (AP).  
The best and second-best results are shown in \textbf{bold} and \underline{underlined}, respectively. Training or tuning-based methods without released training codes are marked with a $^{\dag}$.}
\vspace{-2mm}
\setlength{\tabcolsep}{3pt}
\renewcommand{\arraystretch}{1.1}
\scalebox{0.6}{
\begin{tabular}{l|l|cccccccccc}
\toprule
\multirow{1}{*}{\textbf{Category}} & \multirow{1}{*}{\textbf{Metrics}}
& AnyDoor \cite{chen2024anydoor} & PbE \cite{yang2023paint} & FreeComp. \cite{chen2024freecompose} & FreeCustom \cite{ding2024freecustom} & PrimeComp. \cite{wang2024primecomposer}
& OmniGen \cite{xiao2024omnigen} & GenArt. \cite{wang2024genartist} & UniCom. \cite{wang2025unicombine}
& GPT-4o$^{\dag}$ \cite{chatgpt} & Ours \\
\midrule
\multirow{5}{*}{Automatic}
 & FID $\downarrow$          & 18.57 & 15.91 & 22.55 & 18.57 & 17.48 & 12.13 & 14.52 & 11.55 & \underline{9.98} & \textbf{9.27} \\
 & CLIP-Score $\uparrow$     & 27.65 & 29.03 & 27.56 & 28.43 & 28.31 & \underline{29.77} & 29.11 & 29.28 & 29.35 & \textbf{30.29} \\
 & HOI-Score $\uparrow$      & 25.69 & 38.71 & 22.81 & 45.72 & 32.66 & 62.33 & 51.83 & 58.91 & \underline{75.22} & \textbf{87.39} \\
 & DINO-Score $\uparrow$     & 58.83 & 54.83 & 44.67 & 42.02 & 48.12 & 43.92 & 53.96 & 51.02 & \underline{65.23} & \textbf{78.21} \\
 & SSIM(BG) $\uparrow$         & \underline{90.71} & 88.72 & 86.65 & 43.22 & 85.22 & 82.08 & 57.83 & 88.24 & 47.22 & \textbf{96.57} \\
\midrule
\multirow{3}{*}{User study}
 & IQ $\downarrow$           & 9.72 & 7.47 & 8.20 & 9.13 & 3.23 & \underline{2.63} & 6.22 & 3.93 & 3.10 & \textbf{1.37} \\
 & IH $\downarrow$           & 8.18 & 8.23 & 8.46 & 6.72 & 6.68 & 5.23 & 4.88 & 2.87 & \underline{2.61} & \textbf{1.14} \\
 & AP $\downarrow$           & \underline{2.84} & 5.41 & 6.84 & 7.33 & 6.07 & 4.73 & 6.54 & 8.26 & 5.87 & \textbf{1.11} \\
\bottomrule
\end{tabular}}
\label{tab:sota}
\vspace{3mm}
\end{table*}

\begin{figure}[t]
    \centering
    \begin{overpic}[width=1\textwidth]{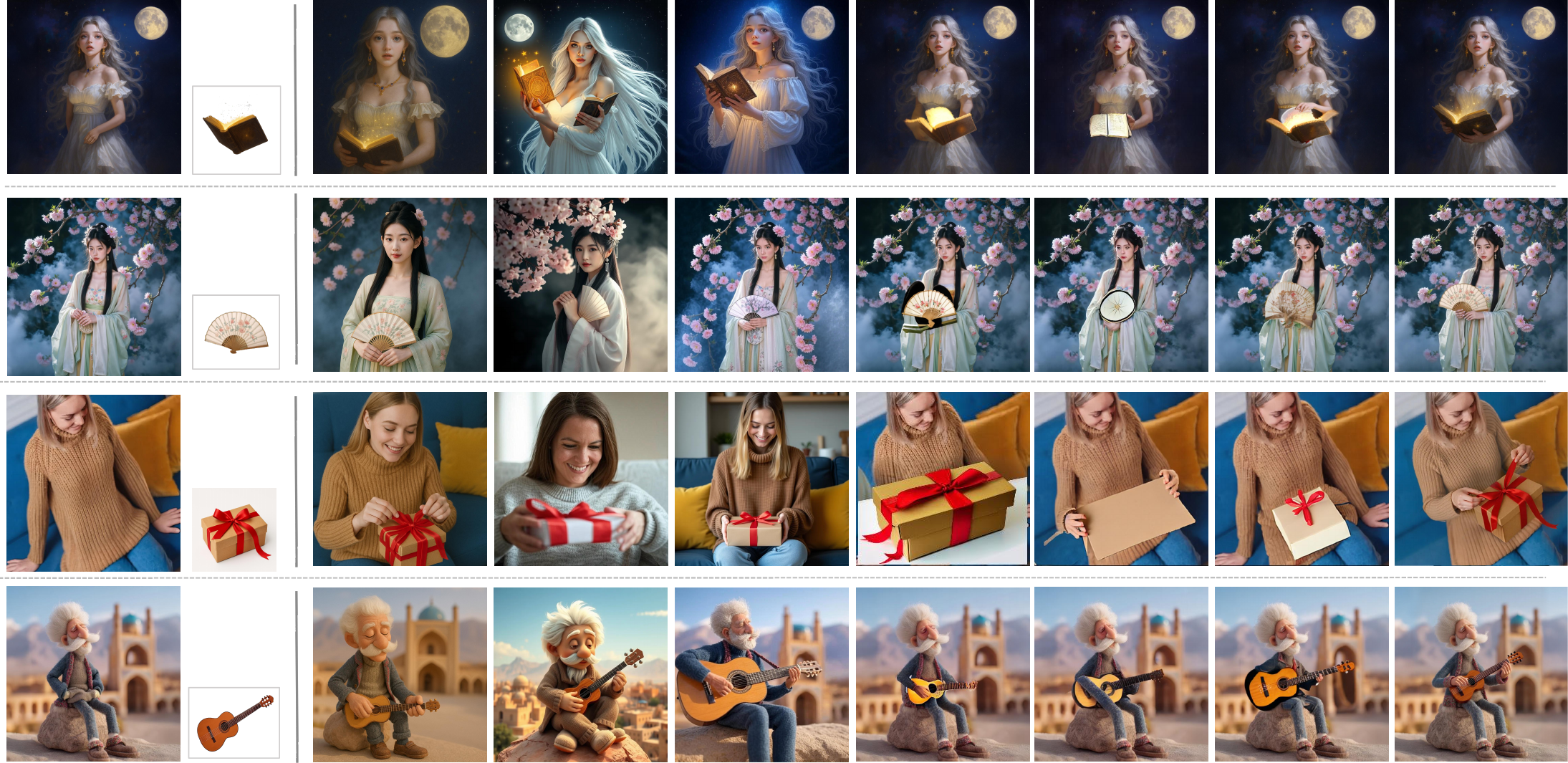}
    \put(2,50){\tiny{(a) Input image \& object}}
    \put(20.5,50){\tiny{(b) GPT-4o\cite{chatgpt}}}
    \put(32.5,50){\tiny{(c) GenArt.\cite{wang2024genartist}}} 
    \put(43,50){\tiny{(d) OmniGen\cite{xiao2024omnigen}}} 
    \put(55,50){\tiny{(e) AnyDoor\cite{chen2024anydoor}}} 
    \put(68,50){\tiny{(f) PbE\cite{yang2023paint}}}    
    \put(78,50){\tiny{(g) UniCom.\cite{wang2025unicombine}}} 
    \put(91.5,50){\tiny{(h) Ours}}
    \end{overpic}
    \vspace{-1mm}
    \caption{Qualitative comparison with six top performing SOTA methods from Table~\ref{tab:sota}. The prompts for the above four examples are: ``A girl is reading a magic book'', ``A woman is holding an ornate folding fan'', ``A woman is opening a gift box'', and ``A puppet-style old man is playing a guitar''.}
    \label{fig:sota}
\end{figure}
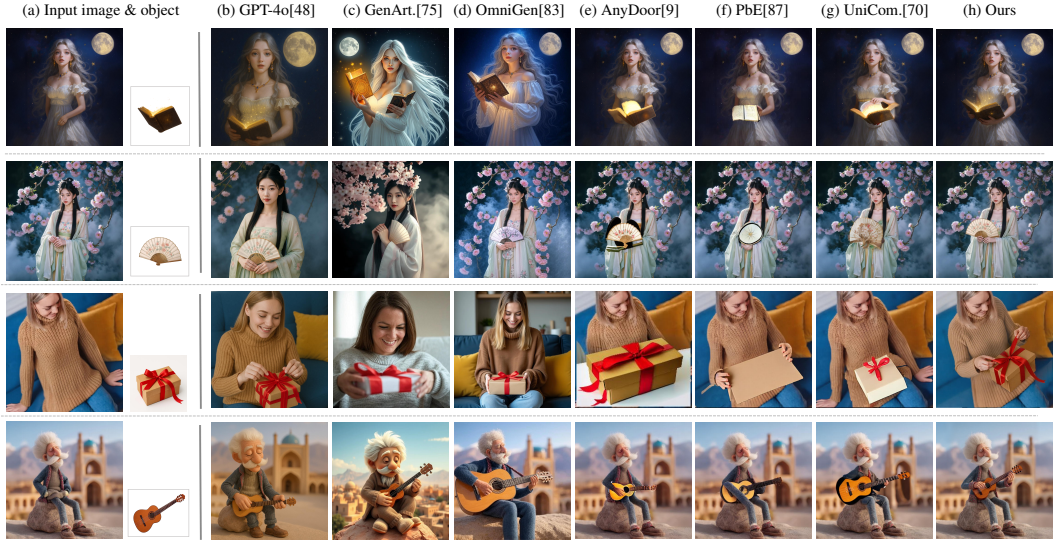

\subsection{Comparison with State-of-the-Art Methods}
\label{subsec:sota}
We compare \name ~with 9 SOTA methods: AnyDoor~\cite{chen2024anydoor}, Paint by Example~\cite{yang2023paint}, FreeCompose~\cite{chen2024freecompose}, FreeCustom~\cite{ding2024freecustom}, OmniGen~\cite{xiao2024omnigen}, GenArtist~\cite{wang2024genartist}, PrimeComposer~\cite{wang2024primecomposer}, UniCombine~\cite{wang2025unicombine} and GPT-4o~\cite{chatgpt}. All methods with public training code are retrained or fine-tuned on our dataset.

\noindent \textbf{Quantitative Comparison.}
Table~\ref{tab:sota} compares the performances of our method against the nine existing methods. The results in the top part of the table show that our method consistently outperforms all these baselines across all evaluation metrics. Specifically, it achieves the highest HOI-Score (87.39), surpassing GPT-4o by 12.17 and OmniGen by 25.06, underscoring its strong ability to model accurate and coherent human–object interactions.
In terms of visual consistency, our method achieves the lowest FID (9.27) and the highest CLIP-Score (30.29), demonstrating superior realism and semantic alignment ability.
Our DINO-Score (78.21) significantly outperforms AnyDoor by 19.38 and GPT-4o by 13.0, indicating improved foreground appearance consistency.
Further, our model produces the most consistent background details with the highest SSIM(BG) score (96.57), outperforming AnyDoor by 5.86.

\noindent \textbf{Qualitative Comparison.} Fig.~\ref{fig:sota} visually compares the results of our method and those of the six top-performing methods from Table~\ref{tab:sota}.
Rows 3-4 of Fig.~\ref{fig:sota}(b) show that although GPT-4o can synthesize plausible human–object interactions, it fails to maintain foreground appearance consistency. Meanwhile, its generated backgrounds exhibit substantial variations, as shown in rows 1-3 of Fig.~\ref{fig:sota}(b). Similar to GPT-4o, GenArtist and OmniGen also suffer from foreground–background inconsistency. In addition, methods in Fig.~\ref{fig:sota}(e-g) produce suboptimal or implausible hand poses. In contrast, our method effectively constrains the generated human poses as well as the shapes/textures of the foreground objects. As a result, the images produced by our method exhibit superior appearance consistency with harmonious human-object interactions.

\noindent \textbf{User Study.} We have also conducted a user study to compare our method with all 9 existing methods. We recruit a total of 75 student participants for the subjective assessment. Each participant is presented with 10 sets of cases, where each set contains an input human image, a foreground object, a text prompt to describe the interaction, and ten randomly shuffled results from \name ~and the 9 competing methods. 
Participants rank the images based on three criteria: image quality (IQ), interaction harmonization (IH), and appearance preservation (AP). We collect ranking scores from all participants and compute the average ranking for each of the three aspects, as shown in the bottom part of Table~\ref{tab:sota}. These results show that our approach ranks first in all three aspects: image quality (1.37), interaction harmonization (1.14), and appearance preservation (1.11), highlighting it being the most preferred method by all participants.

\begin{table}[t]
\centering
\small
\caption{Ablation study on removing one of the key components from our full model (left table) and adding one of the key components to our base model (right table).
$\mathcal{L}_\text{p}$, $\mathcal{L}_\text{b}$, $\mathcal{L}_\text{a}$, and SAAM denote the pose-guided loss, background consistency loss, multi-view appearance loss, and shape-aware attention modulation, respectively. Best performances are marked in \textbf{bold}.}
\vspace{2mm}
\renewcommand\arraystretch{1.0}
\begin{subtable}[t]{0.48\textwidth}
\centering
\scalebox{0.62}{
\begin{tabular}{cccc|ccccc}
\toprule
$\mathcal{L}_\text{p}$ & $\mathcal{L}_\text{b}$ & $\mathcal{L}_\text{a}$ & SAAM & FID $\downarrow$ & CLIP $\uparrow$ & HOI $\uparrow$ & DINO $\uparrow$ & SSIM(BG) $\uparrow$ \\
\midrule
 & \checkmark & \checkmark & \checkmark & 14.24 & 28.05 & 34.42 & 69.32 & 94.91 \\
\checkmark &  & \checkmark & \checkmark & 15.45 & 28.42 & 54.47 & 59.72 & 58.49 \\
\checkmark & \checkmark &  & \checkmark & 13.31 & 29.37 & 67.32 & 46.12 & 95.11 \\
\checkmark & \checkmark & \checkmark &  & 12.48 & 29.10 & 75.23 & 66.52 & 95.28 \\
\checkmark & \checkmark & \checkmark & \checkmark & \textbf{9.27} & \textbf{30.29} & \textbf{87.39} & \textbf{78.21} & \textbf{96.57} \\
\bottomrule
\end{tabular}}
\end{subtable}
\hfill
\begin{subtable}[t]{0.48\textwidth}
\centering
\scalebox{0.62}{
\begin{tabular}{cccc|ccccc}
\toprule
$\mathcal{L}_\text{p}$ & $\mathcal{L}_\text{b}$ & $\mathcal{L}_\text{a}$ & SAAM & FID $\downarrow$ & CLIP $\uparrow$ & HOI $\uparrow$ & DINO $\uparrow$ & SSIM(BG) $\uparrow$ \\
\midrule 
 &  &  &  & 21.25 & 26.14 & 26.76 & 22.19 & 34.91 \\
\checkmark &  &  &  & 15.80 & 26.42 & 47.32 & 30.21 & 53.11 \\
 & \checkmark &  &  & 14.72 & 26.83 & 30.08 & 33.54 & 93.29 \\
 &  & \checkmark &  & 16.02 & 26.71 & 31.09 & 55.81 & 54.29 \\
 &  &  & \checkmark & 16.21 & 26.51 & 29.85 & 42.53 & 57.32 \\
\bottomrule
\end{tabular}}
\end{subtable}
\label{tab:ablation_dual}
\end{table}

\subsection{Ablation Study}

\begin{figure}[t]
\centering
\includegraphics[width=1\linewidth]{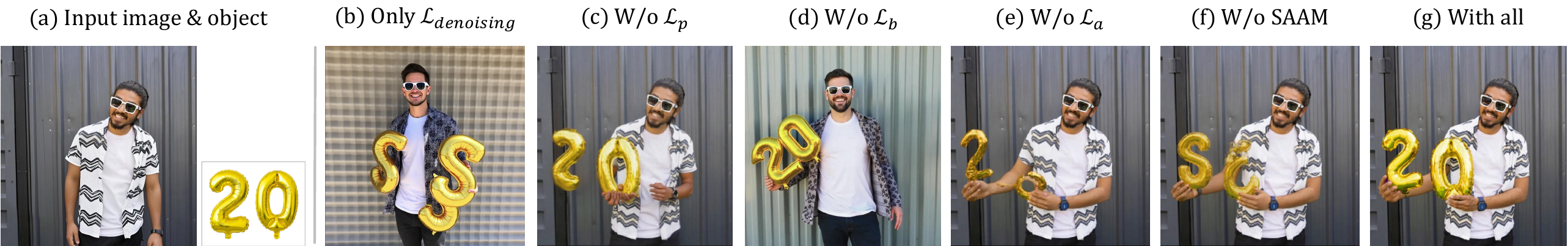}
\caption{Visual comparison of the ablation study in Table~\ref{tab:ablation_dual}.}
\label{fig:abl_loss}
\end{figure}

\noindent \textbf{Component Analysis.} 
We conduct an ablation study on \name{} by systematically removing one key component from our full model (Table~\ref{tab:ablation_dual} (left)) or by adding one key component to our base model (Table~\ref{tab:ablation_dual} (right)).
Fig.~\ref{fig:abl_loss} visualizes some results of the ablation study. Based on these results, we can draw six key conclusions:
%
\ding{182} Pose constraint ($\mathcal{L}_\text{p}$) is essential for ensuring proper human pose generation during interactions. When removed, the result in Fig.~\ref{fig:abl_loss}(c) exhibits a distorted and incongruous interaction, leading to the lowest CLIP and HOI scores shown in row 1 of Table~\ref{tab:ablation_dual} (left). Its absence also lowers the SSIM(BG) score from 96.57 to 94.91, showing a mild but noticeable loss of background consistency.
%
\ding{183} Background consistency loss ($\mathcal{L}_\text{b}$) helps prevent unintended modifications of non-interaction region of the background image. Without it, the person as well as the background scene may undergo significant changes (Fig.~\ref{fig:abl_loss}(d)), resulting in the worst FID score shown in row 2 of Table~\ref{tab:ablation_dual} (left). As a result, the SSIM(BG) score plummets to 58.49, the largest drop among all settings, causing the most severe background degradation.
%
\ding{184} Multi-view appearance loss ($\mathcal{L}_\text{a}$) ensures consistency in the texture/appearance of the foreground object in the generated image. Removing it leads to noticeable color and texture shifts of the object (\eg, the balloons in Fig.~\ref{fig:abl_loss}(e)) and the lowest DINO score shown in row 3 of Table~\ref{tab:ablation_dual} (left). 
\ding{185} Shape-aware attention modulation (SAAM) plays a crucial role in preserving object shape consistency. As shown in row 4 of Table~\ref{tab:ablation_dual} (left), removing SAAM leads to inconsistent shape transformations and appearance variations, with the DINO score dropping significantly from 78.21 to 66.52.
%
\ding{186} Finally, by integrating all key components, our proposed method achieves the best performance, as shown in row 5 of Table~\ref{tab:ablation_dual} (left).
\noindent
\ding{187} Table~\ref{tab:ablation_dual} (right) shows that each component individually enhances a specific aspect of the model. $\mathcal{L}_{p}$ helps improve interaction quality, as reflected in higher HOI and CLIP scores. $\mathcal{L}_{b}$ improves background consistency, evident from the SSIM(BG) score. $\mathcal{L}_{a}$ and SAAM help maintain foreground appearance consistency, leading to improved DINO performances.

\section{Conclusion}

\begin{wrapfigure}{r}{0.32\linewidth}
\centering
\vspace{-4mm}
\includegraphics[width=\linewidth]{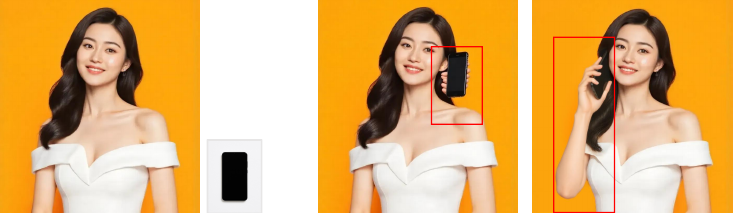}
\put(-132,42){\tiny{Input image \& object}}
\put(-71,42){\tiny{Incorrect $B_r$}}
\put(-32,42){\tiny{Correct $B_r$}}
\caption{An example failure case of \name. The red boxes indicate the interaction regions.}
\vspace{-10pt}
\label{fig:limi}
\end{wrapfigure}

In this paper, we have presented \name, a framework for interaction-aware human-object composition. It leverages MLLM-driven region-based pose guidance (MRPG) for constrained human-object interaction, and detail-consistent appearance preservation (DCAP) for maintaining appearance consistency. To support \name~training, we have also introduced the Interaction-aware Human-Object Composition (IHOC) dataset. Extensive experiments demonstrate that \name{} outperforms existing methods in quantitative, qualitative, and subjective evaluations.

\name{} does have limitations. Although MLLMs correctly identify the interaction region in 91.33\% of the samples in our benchmark, HOIBench, incorrect predictions may still affect the quality of the generated interactions, as shown in Fig.~\ref{fig:limi}. As a future work, we would like to consider incorporating human pose priors into predicting the interaction region.

\bibliography{main}

\begin{thebibliography}{10}

\bibitem{avrahami2023break}
Omri Avrahami, Kfir Aberman, Ohad Fried, Daniel Cohen-Or, and Dani Lischinski.
\newblock Break-a-scene: Extracting multiple concepts from a single image.
\newblock In {\em ACM SIGGRAPH Asia}, pages 1--12, 2023.

\bibitem{batifol2025flux}
Stephen Batifol, Andreas Blattmann, Frederic Boesel, Saksham Consul, Cyril Diagne, Tim Dockhorn, Jack English, Zion English, Patrick Esser, Sumith Kulal, et~al.
\newblock Flux. 1 kontext: Flow matching for in-context image generation and editing in latent space.
\newblock {\em arXiv e-prints}, pages arXiv--2506, 2025.

\bibitem{flux}
{Black Forest Labs}.
\newblock {FLUX.1-dev: A 12B Parameter Rectified Flow Transformer for Text-to-Image Generation}.
\newblock \url{https://huggingface.co/spaces/black-forest-labs/FLUX.1-dev}, 2024.

\bibitem{flux-fill}
{Black Forest Labs}.
\newblock {FLUX.1-Fill-dev: A 12B Parameter Rectified Flow Transformer for Inpainting and Outpainting}.
\newblock \url{https://huggingface.co/black-forest-labs/FLUX.1-Fill-dev}, 2024.

\bibitem{chao2018learning}
Yu-Wei Chao, Yunfan Liu, Xieyang Liu, Huayi Zeng, and Jia Deng.
\newblock Learning to detect human-object interactions.
\newblock In {\em Proceedings of the IEEE Winter Conference on Applications of Computer Vision (WACV)}, pages 381--389, 2018.

\bibitem{chen2024virtualmodel}
Binghui Chen, Chongyang Zhong, Wangmeng Xiang, Yifeng Geng, and Xuansong Xie.
\newblock Virtualmodel: Generating object-id-retentive human-object interaction image by diffusion model for e-commerce marketing.
\newblock {\em arXiv:2405.09985}, 2024.

\bibitem{chen2024mureobjectstitch}
Jiaxuan Chen, Bo~Zhang, Qingdong He, Jinlong Peng, and Li~Niu.
\newblock Mureobjectstitch: Multi-reference image composition.
\newblock {\em arXiv:2411.07462}, 2024.

\bibitem{chen2024zero}
Xi~Chen, Yutong Feng, Mengting Chen, Yiyang Wang, Shilong Zhang, Yu~Liu, Yujun Shen, and Hengshuang Zhao.
\newblock Zero-shot image editing with reference imitation.
\newblock In {\em NeurIPS}, volume~37, pages 84010--84032, 2024.

\bibitem{chen2024anydoor}
Xi~Chen, Lianghua Huang, Yu~Liu, Yujun Shen, Deli Zhao, and Hengshuang Zhao.
\newblock Anydoor: Zero-shot object-level image customization.
\newblock In {\em CVPR}, pages 6593--6602, 2024.

\bibitem{chen2024unireal}
Xi~Chen, Zhifei Zhang, He~Zhang, Yuqian Zhou, Soo~Ye Kim, Qing Liu, Yijun Li, Jianming Zhang, Nanxuan Zhao, Yilin Wang, et~al.
\newblock Unireal: Universal image generation and editing via learning real-world dynamics.
\newblock {\em arXiv:2412.07774}, 2024.

\bibitem{chen2024freecompose}
Zhekai Chen, Wen Wang, Zhen Yang, Zeqing Yuan, Hao Chen, and Chunhua Shen.
\newblock Freecompose: Generic zero-shot image composition with diffusion prior.
\newblock In {\em ECCV}, pages 70--87. Springer, 2024.

\bibitem{deng2025cinema}
Yufan Deng, Xun Guo, Yizhi Wang, Jacob~Zhiyuan Fang, Angtian Wang, Shenghai Yuan, Yiding Yang, Bo~Liu, Haibin Huang, and Chongyang Ma.
\newblock Cinema: Coherent multi-subject video generation via mllm-based guidance.
\newblock {\em arXiv:2503.10391}, 2025.

\bibitem{ding2024freecustom}
Ganggui Ding, Canyu Zhao, Wen Wang, Zhen Yang, Zide Liu, Hao Chen, and Chunhua Shen.
\newblock Freecustom: Tuning-free customized image generation for multi-concept composition.
\newblock In {\em CVPR}, pages 9089--9098, 2024.

\bibitem{gafni2020wish}
Oran Gafni and Lior Wolf.
\newblock Wish you were here: Context-aware human generation.
\newblock In {\em CVPR}, pages 7840--7849, 2020.

\bibitem{gu2023mix}
Yuchao Gu, Xintao Wang, Jay~Zhangjie Wu, Yujun Shi, Yunpeng Chen, Zihan Fan, Wuyou Xiao, Rui Zhao, Shuning Chang, and Weijia Wu.
\newblock Mix-of-show: Decentralized low-rank adaptation for multi-concept customization of diffusion models.
\newblock In {\em NeurIPS}, volume~36, pages 15890--15902, 2023.

\bibitem{he2024affordance}
Jixuan He, Wanhua Li, Ye~Liu, Junsik Kim, Donglai Wei, and Hanspeter Pfister.
\newblock Affordance-aware object insertion via mask-aware dual diffusion.
\newblock {\em arXiv:2412.14462}, 2024.

\bibitem{hessel2021clipscore}
Jack Hessel, Ari Holtzman, Maxwell Forbes, and Yejin Choi.
\newblock Clipscore: A reference-free evaluation metric for image captioning.
\newblock In {\em EMNLP}, 2021.

\bibitem{heusel2017gans}
Martin Heusel, Hubert Ramsauer, Thomas Unterthiner, Bernhard Nessler, and Sepp Hochreiter.
\newblock Gans trained by a two time-scale update rule converge to a local nash equilibrium.
\newblock In {\em NeurIPS}, 2017.

\bibitem{hoe2024interactdiffusion}
Jiun~Tian Hoe, Xudong Jiang, Chee~Seng Chan, Yap-Peng Tan, and Weipeng Hu.
\newblock Interactdiffusion: Interaction control in text-to-image diffusion models.
\newblock In {\em CVPR}, pages 6180--6189, 2024.

\bibitem{hollein2024viewdiff}
Lukas H{\"o}llein, Alja{\v{z}} Bo{\v{z}}i{\v{c}}, Norman M{\"u}ller, David Novotny, Hung-Yu Tseng, Christian Richardt, Michael Zollh{\"o}fer, and Matthias Nie{\ss}ner.
\newblock Viewdiff: 3d-consistent image generation with text-to-image models.
\newblock In {\em CVPR}, pages 5043--5052, 2024.

\bibitem{hou2020visual}
Zhi Hou, Xiaojiang Peng, Yu~Qiao, and Dacheng Tao.
\newblock Visual compositional learning for human-object interaction detection.
\newblock In {\em ECCV}, pages 584--600. Springer, 2020.

\bibitem{hu2022lora}
Edward~J. Hu, Yelong Shen, Phillip Wallis, Zeyuan Allen-Zhu, Yuanzhi Li, Shean Wang, Lu~Wang, and Weizhu Chen.
\newblock Lora: Low-rank adaptation of large language models.
\newblock In {\em ICLR}, 2022.

\bibitem{hu2025personahoi}
Xinting Hu, Haoran Wang, Jan~Eric Lenssen, and Bernt Schiele.
\newblock Personahoi: Effortlessly improving personalized face with human-object interaction generation.
\newblock In {\em CVPR}, 2025.

\bibitem{hua2021exploiting}
Tianyu Hua, Hongdong Zheng, Yalong Bai, Wei Zhang, Xiao-Ping Zhang, and Tao Mei.
\newblock Exploiting relationship for complex-scene image generation.
\newblock In {\em AAAI}, volume~35, pages 1584--1592, 2021.

\bibitem{huang2025dreamfuse}
Junjia Huang, Pengxiang Yan, Jiyang Liu, Jie Wu, Zhao Wang, Yitong Wang, Liang Lin, and Guanbin Li.
\newblock Dreamfuse: Adaptive image fusion with diffusion transformer.
\newblock {\em arXiv:2504.08291}, 2025.

\bibitem{huang2024learning}
Siteng Huang, Biao Gong, Yutong Feng, Xi~Chen, Yuqian Fu, Yu~Liu, and Donglin Wang.
\newblock Learning disentangled identifiers for action-customized text-to-image generation.
\newblock In {\em CVPR}, pages 7797--7806, 2024.

\bibitem{huang2024reversion}
Ziqi Huang, Tianxing Wu, Yuming Jiang, Kelvin~CK Chan, and Ziwei Liu.
\newblock Reversion: Diffusion-based relation inversion from images.
\newblock In {\em SIGGRAPH Asia}, pages 1--11, 2024.

\bibitem{huang2025hunyuanvideo}
Ziyao Huang, Zixiang Zhou, Juan Cao, Yifeng Ma, Yi~Chen, Zejing Rao, Zhiyong Xu, Hongmei Wang, Qin Lin, Yuan Zhou, et~al.
\newblock Hunyuanvideo-homa: Generic human-object interaction in multimodal driven human animation.
\newblock {\em arXiv preprint arXiv:2506.08797}, 2025.

\bibitem{jiang2024record}
Jian-Yu Jiang-Lin, Kang-Yang Huang, Ling Lo, Yi-Ning Huang, Terence Lin, Jhih-Ciang Wu, Hong-Han Shuai, and Wen-Huang Cheng.
\newblock Record: Reasoning and correcting diffusion for hoi generation.
\newblock In {\em ACM MM}, pages 9465--9474, 2024.

\bibitem{kim2023dense}
Yunji Kim, Jiyoung Lee, Jin-Hwa Kim, Jung-Woo Ha, and Jun-Yan Zhu.
\newblock Dense text-to-image generation with attention modulation.
\newblock In {\em ICCV}, pages 7701--7711, 2023.

\bibitem{kingma2013auto}
Diederik~P Kingma and Max Welling.
\newblock Auto-encoding variational bayes.
\newblock {\em arXiv:1312.6114}, 2013.

\bibitem{kong2024omg}
Zhe Kong, Yong Zhang, Tianyu Yang, Tao Wang, Kaihao Zhang, Bizhu Wu, Guanying Chen, Wei Liu, and Wenhan Luo.
\newblock Omg: Occlusion-friendly personalized multi-concept generation in diffusion models.
\newblock In {\em ECCV}, pages 253--270. Springer, 2024.

\bibitem{kumari2023multi}
Nupur Kumari, Bingliang Zhang, Richard Zhang, Eli Shechtman, and Jun-Yan Zhu.
\newblock Multi-concept customization of text-to-image diffusion.
\newblock In {\em CVPR}, pages 1931--1941, 2023.

\bibitem{li2024textit}
Lingxiao Li, Kaixiong Gong, Wei-Hong Li, Tao Chen, Xiaojun Yuan, and Xiangyu Yue.
\newblock Bifrost: 3d-aware image compositing with language instructions.
\newblock In {\em NeurIPS}, volume~37, pages 129480--129506, 2024.

\bibitem{li2024tuning}
Pengzhi Li, Qiang Nie, Ying Chen, Xi~Jiang, Kai Wu, Yuhuan Lin, Yong Liu, Jinlong Peng, Chengjie Wang, and Feng Zheng.
\newblock Tuning-free image customization with image and text guidance.
\newblock In {\em ECCV}, pages 233--250. Springer, 2024.

\bibitem{li2023gligen}
Yuheng Li, Haotian Liu, Qingyang Wu, Fangzhou Mu, Jianwei Yang, Jianfeng Gao, Chunyuan Li, and Yong~Jae Lee.
\newblock Gligen: Open-set grounded text-to-image generation.
\newblock In {\em CVPR}, pages 22511--22521, 2023.

\bibitem{liang2025vodiff}
Dong Liang, Jinyuan Jia, Yuhao Liu, Zhanghan Ke, Hongbo Fu, and Rynson~WH Lau.
\newblock Vodiff: Controlling object visibility order in text-to-image generation.
\newblock In {\em Proceedings of the Computer Vision and Pattern Recognition Conference}, pages 18379--18389, 2025.

\bibitem{liang2025movie}
Feng Liang, Haoyu Ma, Zecheng He, Tingbo Hou, Ji~Hou, Kunpeng Li, Xiaoliang Dai, Felix Juefei-Xu, Samaneh Azadi, Animesh Sinha, et~al.
\newblock Movie weaver: Tuning-free multi-concept video personalization with anchored prompts.
\newblock {\em arXiv:2502.07802}, 2025.

\bibitem{lin2024non}
Wang Lin, Jingyuan Chen, Jiaxin Shi, Yichen Zhu, Chen Liang, Junzhong Miao, Tao Jin, Zhou Zhao, Fei Wu, Shuicheng Yan, et~al.
\newblock Non-confusing generation of customized concepts in diffusion models.
\newblock {\em arXiv:2405.06914}, 2024.

\bibitem{lin2025multitwine}
Zhe Lin, Zhifei Zhang, He~Zhang, Andrew Gilbert, John~Philip Collomosse, and Soo~Ye Kim.
\newblock Multitwine: Multi-object compositing with text and layout control.
\newblock In {\em CVPR}, 2025.

\bibitem{liu2023grounding}
Shilong Liu, Zhaoyang Zeng, Tianhe Ren, Feng Li, Hao Zhang, Jie Yang, Chunyuan Li, Jianwei Yang, Hang Su, Jun Zhu, et~al.
\newblock Grounding dino: Marrying dino with grounded pre-training for open-set object detection.
\newblock {\em arXiv:2303.05499}, 2023.

\bibitem{liu2025step1x}
Shiyu Liu, Yucheng Han, Peng Xing, Fukun Yin, Rui Wang, Wei Cheng, Jiaqi Liao, Yingming Wang, Honghao Fu, Chunrui Han, et~al.
\newblock Step1x-edit: A practical framework for general image editing.
\newblock {\em arXiv:2504.17761}, 2025.

\bibitem{liu2023cones}
Zhiheng Liu, Yifei Zhang, Yujun Shen, Kecheng Zheng, Kai Zhu, Ruili Feng, Yu~Liu, Deli Zhao, Jingren Zhou, and Yang Cao.
\newblock Cones 2: Customizable image synthesis with multiple subjects.
\newblock In {\em NeurIPS}, pages 57500--57519, 2023.

\bibitem{liu2023customizable}
Zhiheng Liu, Yifei Zhang, Yujun Shen, Kecheng Zheng, Kai Zhu, Ruili Feng, Yu~Liu, Deli Zhao, Jingren Zhou, and Yang Cao.
\newblock Customizable image synthesis with multiple subjects.
\newblock In {\em NeurIPS}, volume~36, pages 57500--57519, 2023.

\bibitem{lu2023tf}
Shilin Lu, Yanzhu Liu, and Adams Wai-Kin Kong.
\newblock Tf-icon: Diffusion-based training-free cross-domain image composition.
\newblock In {\em ICCV}, pages 2294--2305, 2023.

\bibitem{midjourney2025}
MidJourney.
\newblock Midjourney official website, 2025.

\bibitem{navard2024knobgen}
Pouyan Navard, Amin~Karimi Monsefi, Mengxi Zhou, Wei-Lun Chao, Alper Yilmaz, and Rajiv Ramnath.
\newblock Knobgen: Controlling the sophistication of artwork in sketch-based diffusion models.
\newblock {\em arXiv:2410.01595}, 2024.

\bibitem{chatgpt}
{OpenAI}.
\newblock {ChatGPT (model 4o)}.
\newblock \url{https://chat.openai.com/}, 2025.

\bibitem{oquabdinov2}
Maxime Oquab, Timoth{\'e}e Darcet, Th{\'e}o Moutakanni, Huy~V Vo, Marc Szafraniec, Vasil Khalidov, Pierre Fernandez, Daniel HAZIZA, Francisco Massa, Alaaeldin El-Nouby, et~al.
\newblock Dinov2: Learning robust visual features without supervision.
\newblock {\em TMLR}.

\bibitem{parihar2024text2place}
Rishubh Parihar, Harsh Gupta, Sachidanand VS, and R~Venkatesh Babu.
\newblock Text2place: Affordance-aware text guided human placement.
\newblock In {\em ECCV}, pages 57--77. Springer, 2024.

\bibitem{parmar2023zero}
Gaurav Parmar, Krishna Kumar~Singh, Richard Zhang, Yijun Li, Jingwan Lu, and Jun-Yan Zhu.
\newblock Zero-shot image-to-image translation.
\newblock In {\em ACM SIGGRAPH}, pages 1--11, 2023.

\bibitem{patellambda}
Maitreya Patel, Sangmin Jung, Chitta Baral, and Yezhou Yang.
\newblock $\lambda$-eclipse: Multi-concept personalized text-to-image diffusion models by leveraging clip latent space.
\newblock {\em TMLR}, 2024.

\bibitem{peebles2023scalable}
William Peebles and Saining Xie.
\newblock Scalable diffusion models with transformers.
\newblock In {\em ICCV}, pages 4195--4205, 2023.

\bibitem{pham2024tale}
Kien~T Pham, Jingye Chen, and Qifeng Chen.
\newblock Tale: Training-free cross-domain image composition via adaptive latent manipulation and energy-guided optimization.
\newblock In {\em ACM MM}, pages 3160--3169, 2024.

\bibitem{radford2021learning}
Alec Radford, Jong~Wook Kim, Chris Hallacy, Aditya Ramesh, Gabriel Goh, Sandhini Agarwal, Girish Sastry, Amanda Askell, Pamela Mishkin, Jack Clark, et~al.
\newblock Learning transferable visual models from natural language supervision.
\newblock In {\em ICML}, pages 8748--8763, 2021.

\bibitem{ravi2024sam}
Nikhila Ravi, Valentin Gabeur, Yuan-Ting Hu, Ronghang Hu, Chaitanya Ryali, Tengyu Ma, Haitham Khedr, Roman R{\"a}dle, Chloe Rolland, Laura Gustafson, et~al.
\newblock Sam 2: Segment anything in images and videos.
\newblock {\em arXiv:2408.00714}, 2024.

\bibitem{ren2024relightful}
Mengwei Ren, Wei Xiong, Jae~Shin Yoon, Zhixin Shu, Jianming Zhang, HyunJoon Jung, Guido Gerig, and He~Zhang.
\newblock Relightful harmonization: Lighting-aware portrait background replacement.
\newblock In {\em CVPR}, pages 6452--6462, 2024.

\bibitem{ruiz2024magic}
Nataniel Ruiz, Yuanzhen Li, Neal Wadhwa, Yael Pritch, Michael Rubinstein, David~E Jacobs, and Shlomi Fruchter.
\newblock Magic insert: Style-aware drag-and-drop.
\newblock {\em arXiv:2407.02489}, 2024.

\bibitem{safaee2024clic}
Mehdi Safaee, Aryan Mikaeili, Or~Patashnik, Daniel Cohen-Or, and Ali Mahdavi-Amiri.
\newblock Clic: Concept learning in context.
\newblock In {\em CVPR}, pages 6924--6933, 2024.

\bibitem{shi2025dreamrelation}
Qingyu Shi, Lu~Qi, Jianzong Wu, Jinbin Bai, Jingbo Wang, Yunhai Tong, and Xiangtai Li.
\newblock Dreamrelation: Bridging customization and relation generation.
\newblock In {\em CVPR}, 2025.

\bibitem{shi2023zero123++}
Ruoxi Shi, Hansheng Chen, Zhuoyang Zhang, Minghua Liu, Chao Xu, Xinyue Wei, Linghao Chen, Chong Zeng, and Hao Su.
\newblock Zero123++: a single image to consistent multi-view diffusion base model.
\newblock {\em arXiv:2310.15110}, 2023.

\bibitem{song2025insert}
Wensong Song, Hong Jiang, Zongxing Yang, Ruijie Quan, and Yi~Yang.
\newblock Insert anything: Image insertion via in-context editing in dit.
\newblock {\em arXiv:2504.15009}, 2025.

\bibitem{song2023objectstitch}
Yizhi Song, Zhifei Zhang, Zhe Lin, Scott Cohen, Brian Price, Jianming Zhang, Soo~Ye Kim, and Daniel Aliaga.
\newblock Objectstitch: Object compositing with diffusion model.
\newblock In {\em CVPR}, pages 18310--18319, 2023.

\bibitem{song2024imprint}
Yizhi Song, Zhifei Zhang, Zhe Lin, Scott Cohen, Brian Price, Jianming Zhang, Soo~Ye Kim, He~Zhang, Wei Xiong, and Daniel Aliaga.
\newblock Imprint: Generative object compositing by learning identity-preserving representation.
\newblock In {\em CVPR}, pages 8048--8058, 2024.

\bibitem{tao2024motioncom}
Weijing Tao, Xiaofeng Yang, Miaomiao Cui, and Guosheng Lin.
\newblock Motioncom: Automatic and motion-aware image composition with llm and video diffusion prior.
\newblock {\em arXiv:2409.10090}, 2024.

\bibitem{tarres2024thinking}
Gemma~Canet Tarr{\'e}s, Zhe Lin, Zhifei Zhang, Jianming Zhang, Yizhi Song, Dan Ruta, Andrew Gilbert, John Collomosse, and Soo~Ye Kim.
\newblock Thinking outside the bbox: Unconstrained generative object compositing.
\newblock {\em arXiv:2409.04559}, 2024.

\bibitem{tewel2023key}
Yoad Tewel, Rinon Gal, Gal Chechik, and Yuval Atzmon.
\newblock Key-locked rank one editing for text-to-image personalization.
\newblock In {\em ACM SIGGRAPH}, pages 1--11, 2023.

\bibitem{tian2025mige}
Xueyun Tian, Wei Li, Bingbing Xu, Yige Yuan, Yuanzhuo Wang, and Huawei Shen.
\newblock Mige: A unified framework for multimodal instruction-based image generation and editing.
\newblock {\em arXiv:2502.21291}, 2025.

\bibitem{voleti2024sv3d}
Vikram Voleti, Chun-Han Yao, Mark Boss, Adam Letts, David Pankratz, Dmitry Tochilkin, Christian Laforte, Robin Rombach, and Varun Jampani.
\newblock Sv3d: Novel multi-view synthesis and 3d generation from a single image using latent video diffusion.
\newblock In {\em ECCV}, pages 439--457. Springer, 2024.

\bibitem{wang2025unicombine}
Haoxuan Wang, Jinlong Peng, Qingdong He, Hao Yang, Ying Jin, Jiafu Wu, Xiaobin Hu, Yanjie Pan, Zhenye Gan, Mingmin Chi, et~al.
\newblock Unicombine: Unified multi-conditional combination with diffusion transformer.
\newblock {\em arXiv:2503.09277}, 2025.

\bibitem{wang2025dreamactor}
Lizhen Wang, Zhurong Xia, Tianshu Hu, Pengrui Wang, Pengfei Wang, Zerong Zheng, and Ming Zhou.
\newblock Dreamactor-h1: High-fidelity human-product demonstration video generation via motion-designed diffusion transformers.
\newblock {\em arXiv preprint arXiv:2506.10568}, 2025.

\bibitem{wangms}
Xierui Wang, Siming Fu, Qihan Huang, Wanggui He, and Hao Jiang.
\newblock Ms-diffusion: Multi-subject zero-shot image personalization with layout guidance.
\newblock In {\em ICLR}, 2025.

\bibitem{wang2024magicface}
Yibin Wang, Weizhong Zhang, and Cheng Jin.
\newblock Magicface: Training-free universal-style human image customized synthesis.
\newblock {\em arXiv:2408.07433}, 2024.

\bibitem{wang2024primecomposer}
Yibin Wang, Weizhong Zhang, Jianwei Zheng, and Cheng Jin.
\newblock Primecomposer: Faster progressively combined diffusion for image composition with attention steering.
\newblock In {\em ACM MM}, pages 10824--10832, 2024.

\bibitem{wang2024genartist}
Zhenyu Wang, Aoxue Li, Zhenguo Li, and Xihui Liu.
\newblock Genartist: Multimodal llm as an agent for unified image generation and editing.
\newblock In {\em NeurIPS}, volume~37, pages 128374--128395, 2024.

\bibitem{wang2004ssim}
Zhou Wang, Alan~C. Bovik, Hamid~R. Sheikh, and Eero~P. Simoncelli.
\newblock Image quality assessment: From error visibility to structural similarity.
\newblock {\em IEEE TIP}, 13(4):600--612, 2004.

\bibitem{wei2025dreamrelation}
Yujie Wei, Shiwei Zhang, Hangjie Yuan, Biao Gong, Longxiang Tang, Xiang Wang, Haonan Qiu, Hengjia Li, Shuai Tan, Yingya Zhang, et~al.
\newblock Dreamrelation: Relation-centric video customization.
\newblock {\em arXiv:2503.07602}, 2025.

\bibitem{winter2024objectdrop}
Daniel Winter, Matan Cohen, Shlomi Fruchter, Yael Pritch, Alex Rav-Acha, and Yedid Hoshen.
\newblock Objectdrop: Bootstrapping counterfactuals for photorealistic object removal and insertion.
\newblock In {\em ECCV}, pages 112--129. Springer, 2024.

\bibitem{winter2024objectmate}
Daniel Winter, Asaf Shul, Matan Cohen, Dana Berman, Yael Pritch, Alex Rav-Acha, and Yedid Hoshen.
\newblock Objectmate: A recurrence prior for object insertion and subject-driven generation.
\newblock {\em arXiv:2412.08645}, 2024.

\bibitem{woo2025flipconcept}
Young~Beom Woo and Sun~Eung Kim.
\newblock Flipconcept: Tuning-free multi-concept personalization for text-to-image generation.
\newblock {\em arXiv:2502.15203}, 2025.

\bibitem{xai2025grok3}
xAI.
\newblock Grok 3: The age of reasoning agents.
\newblock \url{https://x.ai/news/grok-3}, 2025.

\bibitem{xiao2024fastcomposer}
Guangxuan Xiao, Tianwei Yin, William~T Freeman, Fr{\'e}do Durand, and Song Han.
\newblock Fastcomposer: Tuning-free multi-subject image generation with localized attention.
\newblock {\em IJCV}, pages 1--20, 2024.

\bibitem{xiao2024omnigen}
Shitao Xiao, Yueze Wang, Junjie Zhou, Huaying Yuan, Xingrun Xing, Ruiran Yan, Chaofan Li, Shuting Wang, Tiejun Huang, and Zheng Liu.
\newblock Omnigen: Unified image generation.
\newblock In {\em CVPR}, 2025.

\bibitem{xu2024anchorcrafter}
Ziyi Xu, Ziyao Huang, Juan Cao, Yong Zhang, Xiaodong Cun, Qing Shuai, Yuchen Wang, Linchao Bao, Jintao Li, and Fan Tang.
\newblock Anchorcrafter: Animate cyberanchors saling your products via human-object interacting video generation.
\newblock {\em arXiv:2411.17383}, 2024.

\bibitem{xue2022dccf}
Ben Xue, Shenghui Ran, Quan Chen, Rongfei Jia, Binqiang Zhao, and Xing Tang.
\newblock Dccf: Deep comprehensible color filter learning framework for high-resolution image harmonization.
\newblock In {\em ECCV}, pages 300--316. Springer, 2022.

\bibitem{xue2024hoi}
Zihui~Sherry Xue, Romy Luo, Changan Chen, and Kristen Grauman.
\newblock Hoi-swap: Swapping objects in videos with hand-object interaction awareness.
\newblock In {\em NeurIPS}, volume~37, pages 77132--77164, 2024.

\bibitem{yang2023paint}
Binxin Yang, Shuyang Gu, Bo~Zhang, Ting Zhang, Xuejin Chen, Xiaoyan Sun, Dong Chen, and Fang Wen.
\newblock Paint by example: Exemplar-based image editing with diffusion models.
\newblock In {\em CVPR}, pages 18381--18391, 2023.

\bibitem{yang2024person}
ChangHee Yang, ChanHee Kang, Kyeongbo Kong, Hanni Oh, and Suk-Ju Kang.
\newblock Person in place: Generating associative skeleton-guidance maps for human-object interaction image editing.
\newblock In {\em CVPR}, pages 8164--8175, 2024.

\bibitem{yang2023effective}
Zhendong Yang, Ailing Zeng, Chun Yuan, and Yu~Li.
\newblock Effective whole-body pose estimation with two-stages distillation.
\newblock In {\em ICCV}, pages 4210--4220, 2023.

\bibitem{yao2025freegraftor}
Zebin Yao, Lei Ren, Huixing Jiang, Chen Wei, Xiaojie Wang, Ruifan Li, and Fangxiang Feng.
\newblock Freegraftor: Training-free cross-image feature grafting for subject-driven text-to-image generation.
\newblock {\em arXiv:2504.15958}, 2025.

\bibitem{ye2023affordance}
Yufei Ye, Xueting Li, Abhinav Gupta, Shalini De~Mello, Stan Birchfield, Jiaming Song, Shubham Tulsiani, and Sifei Liu.
\newblock Affordance diffusion: Synthesizing hand-object interactions.
\newblock In {\em CVPR}, pages 22479--22489, 2023.

\bibitem{yu2025omnipaint}
Yongsheng Yu, Ziyun Zeng, Haitian Zheng, and Jiebo Luo.
\newblock Omnipaint: Mastering object-oriented editing via disentangled insertion-removal inpainting.
\newblock {\em arXiv:2503.08677}, 2025.

\bibitem{zhang2023controlcom}
Bo~Zhang, Yuxuan Duan, Jun Lan, Yan Hong, Huijia Zhu, Weiqiang Wang, and Li~Niu.
\newblock Controlcom: Controllable image composition using diffusion model.
\newblock {\em arXiv:2308.10040}, 2023.

\bibitem{zhang2022efficient}
Frederic~Z Zhang, Dylan Campbell, and Stephen Gould.
\newblock Efficient two-stage detection of human-object interactions with a novel unary-pairwise transformer.
\newblock In {\em CVPR}, pages 20104--20112, 2022.

\bibitem{zhang2023adding}
Lvmin Zhang, Anyi Rao, and Maneesh Agrawala.
\newblock Adding conditional control to text-to-image diffusion models.
\newblock In {\em ICCV}, pages 3836--3847, 2023.

\bibitem{zhang2023motioncrafter}
Yuxin Zhang, Fan Tang, Nisha Huang, Haibin Huang, Chongyang Ma, Weiming Dong, and Changsheng Xu.
\newblock Motioncrafter: One-shot motion customization of diffusion models.
\newblock {\em arXiv:2312.05288}, 2023.

\bibitem{zhang2024ssr}
Yuxuan Zhang, Yiren Song, Jiaming Liu, Rui Wang, Jinpeng Yu, Hao Tang, Huaxia Li, Xu~Tang, Yao Hu, Han Pan, et~al.
\newblock Ssr-encoder: Encoding selective subject representation for subject-driven generation.
\newblock In {\em CVPR}, pages 8069--8078, 2024.

\end{thebibliography}
\bibliographystyle{plain}

\appendix

\clearpage 

\begin{center}
\Large \textbf{HOComp: Interaction-Aware Human-Object Composition} \\[3ex]
\end{center}

\section*{Appendix}
\addcontentsline{toc}{section}{Appendix} 

\setcounter{section}{0}  
\section{Overview}

In this appendix, we provide additional implementation details, ablation analyses, and extended evaluations to further support and expand upon the findings presented in the main paper.

Specifically, we address the following key aspects in our appendix:
(1) Presenting detailed statistical analyses and the construction procedure of our \textit{IHOC} dataset (\secref{sec:dataset});
(2) Offering additional clarifications on our approach, including experimental configurations and supplementary ablation analyses  (Sec.~\ref{supp:attention}-~\ref{supp:ablation}); 
(3) Presenting additional experiments to validate our method, including further comparisons with state-of-the-art approaches and more results of our method (Sec.~\ref{supp:compare_MLLM}-~\ref{supp:more_results}). 

\section{Extended Details on \textit{IHOC} dataset}
\label{sec:dataset}

\subsection{Dataset Construction}

\begin{figure}[h]
    \centering
    \begin{overpic}[width=1\textwidth]{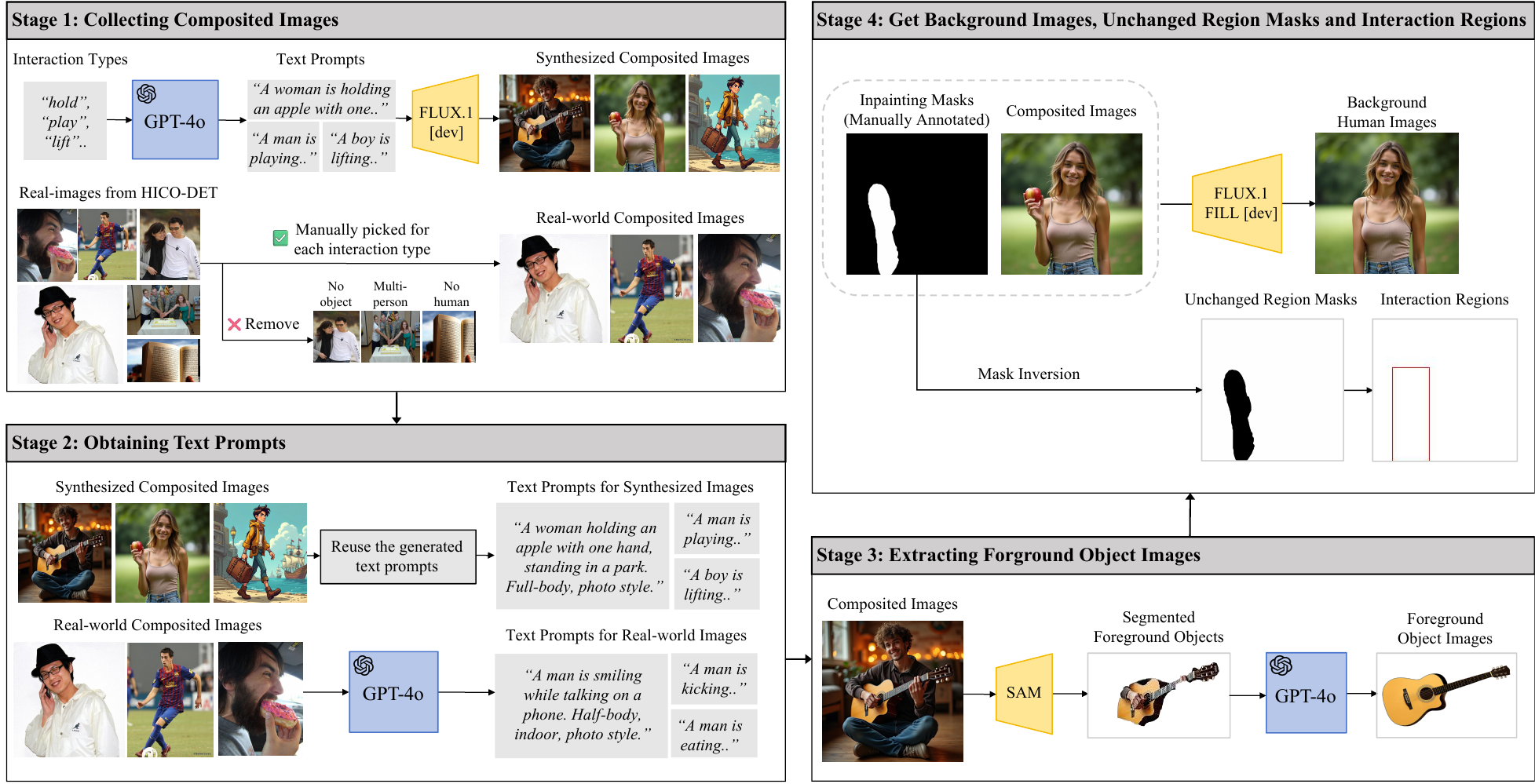}
    \end{overpic}
    \caption{Overview of the construction process of our \textit{Interaction-aware Human-Object Composition (IHOC) dataset}. It involves four stages: (1) collecting synthesized and real-world composited images, (2) obtaining corresponding text prompts, (3) extracting foreground object images, and (4) getting background human images, unchanged region masks, and interaction regions.}
    \label{fig:dataset_con}
\vspace{0mm}
\end{figure}

In Sec. 3.4 of the main paper, we briefly discuss our \textit{Interaction-aware Human-Object Composition (IHOC) dataset}, which includes six components: (1) background human images (without the object); (2) foreground object images; (3) composited images with harmonious interactions and consistent appearances; (4) text prompts describing the interaction type; (5) interaction regions; and (6) unchanged region masks to indicate unaffected background areas. As shown in \figref{fig:dataset_con}, our IHOC dataset construction comprises four stages. 

\textbf{Stage 1: Collecting synthesized and real composited images.}
To ensure data diversity, we adopt the 117 human-object interaction categories from HICO-DET\cite{hou2020visual}, comprising both real and synthetic samples. For real images, we manually selected 50 images per category, resulting in a total of 5,850 from HICO-DET, excluding those that (1) contain multiple people, (2) lack clearly visible humans, or (3) lack clearly visible objects, which impair recognizability. The final set emphasizes diversity in object type, scale, and human pose across scenes. For synthetic images, we use GPT-4o to generate 50 text prompts per category and synthesize 5,850 samples using FLUX.1 [dev]\cite{flux}. These images complement the real data by introducing broader variations in human appearance, pose, viewpoint, and visual style (e.g., cartoons, sketches). In total, we collect 11,700 composited images.

\textbf{Stage 2: Generating text prompts.}  
For real images, we use GPT-4o to generate descriptive prompts. For synthetic images, we reuse the prompts originally used for generation.

\textbf{Stage 3: Extracting foreground objects.}  
We segment foreground objects from composited images using SAM~\cite{ravi2024sam}. To address occlusions caused by human-object interactions, GPT-4o infers and fills missing regions, producing complete and visually consistent objects.

\textbf{Stage 4: Getting background images, unchanged region masks, and interaction regions.}  
We manually annotate inpainting masks and use FLUX.1 FILL [dev]~\cite{flux-fill} to remove interacting objects and reconstruct plausible human poses without interactions. The inpainting masks define interaction-affected regions; their inverse yields the unchanged region masks. Interaction regions are computed by extracting the minimal bounding box of the interaction area within the unchanged region mask.

\subsection{Dataset Statistics}

\begin{figure}[h]
    \centering
    \begin{overpic}[width=1\textwidth]{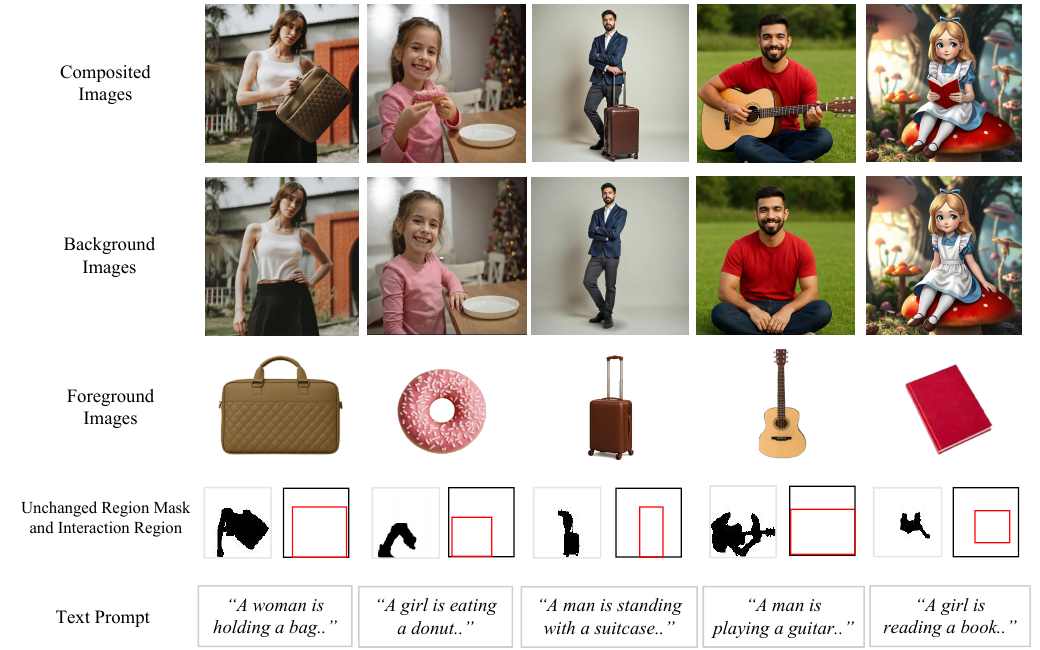}
    \end{overpic}
    \caption{Visualization of our Interaction-aware Human-Object Composition (IHOC) Dataset.
     }
    \label{fig:dataset_vis}
\vspace{0mm}
\end{figure}

As shown in \figref{fig:dataset_vis}, our dataset consists of six components: (1) background human images (without the object); (2) foreground object images; (3) composited images with harmonious interactions and consistent appearances; (4) unchanged region masks to indicate unaffected background areas; (5) interaction regions and (6) text prompts describing the interaction type;

\begin{figure}[t]
    \centering
    \begin{overpic}[width=1\textwidth]{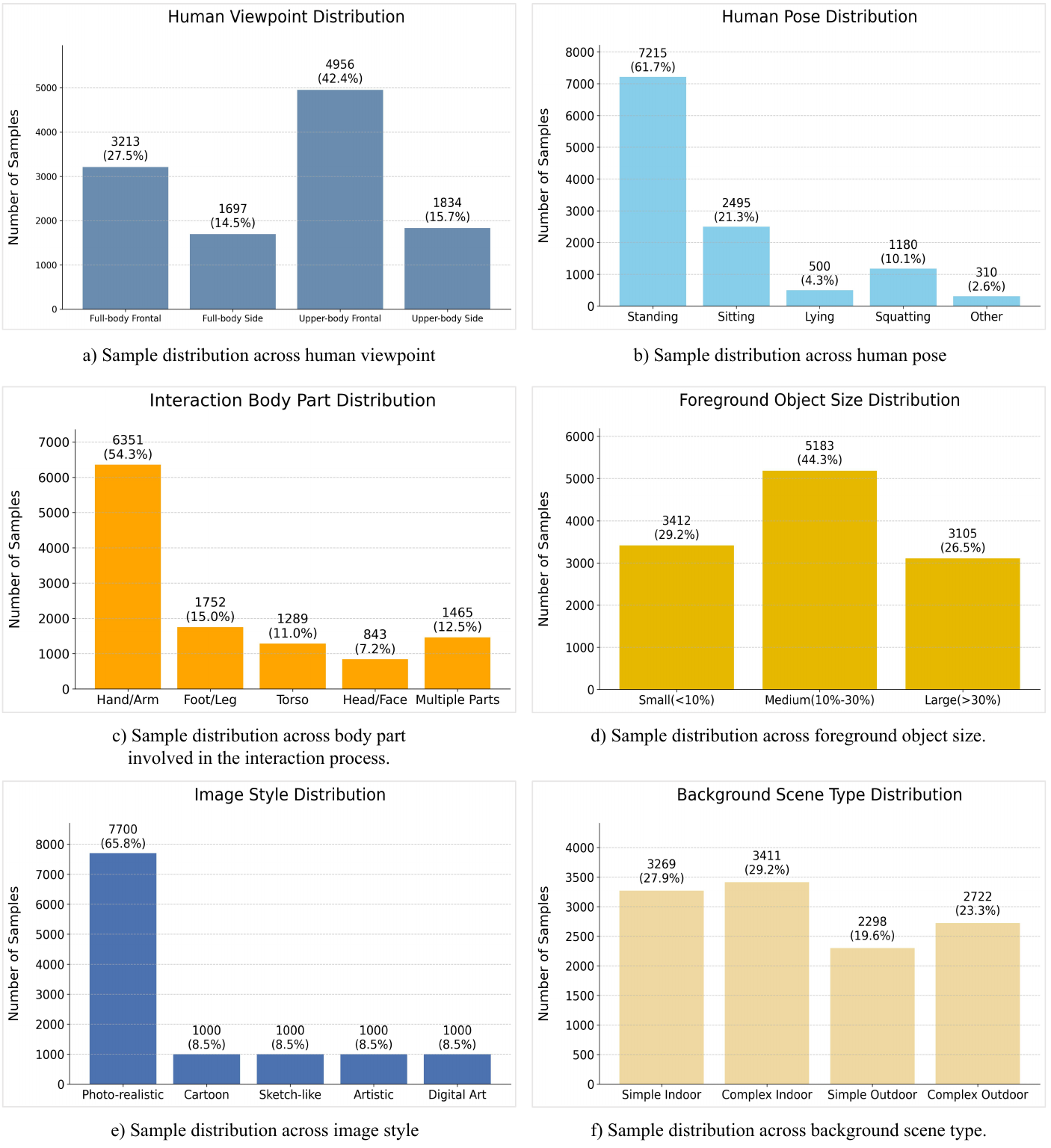}
    \end{overpic}
    \vspace{-2mm}
    \caption{
    Statistical analysis of our \textit{Interaction-aware Human-Object Composition (IHOC) dataset} across six dimensions: (a) human viewpoint, (b) human pose, (c) interaction body part, (d) foreground object size, (e) image style, and (f) background scene type. These statistics demonstrate the dataset's diversity in visual appearance, interaction types, and contextual complexity.
    }
    \label{fig:dataset_sta}
\vspace{0mm}
\end{figure}

Our dataset consists of 11,700 composited images, with half sourced from real-world data and the other half generated synthetically. Our dataset comprises a total of 117 types of interaction types and 342 distinct foreground object categories. To highlight the diversity of our dataset, we analyze its statistical properties across six dimensions, as illustrated in \figref{fig:dataset_sta}(a–f):

\textbf{(1) Human Viewpoint:} Our dataset includes four distinct human viewpoints, categorized by body visibility and camera angle: full-body frontal, full-body side, upper-body frontal, and upper-body side (see Fig.~\ref{fig:dataset_sta}(a)). Upper-body frontal is the most common (42.4\%), followed by full-body frontal (27.5\%), upper-body side (15.7\%), and full-body side (14.5\%). This distribution is reasonable, as frontal views typically support a wider range of interaction types and are more frequently used in practice.

\textbf{(2) Human Pose:} Our dataset covers five major categories of human pose: standing, sitting, lying, squatting, and other (\eg, jumping on a skateboard) (see Fig.~\ref{fig:dataset_sta}(b)). Standing is the most prevalent (61.7\%), followed by sitting (21.3\%), squatting (10.1\%), lying (4.3\%), and other (2.6\%). This distribution demonstrates that our dataset includes both common and less frequent poses.

\textbf{(3) Interaction Body Part:} We categorize the interactions in our dataset into five body regions based on which part of the body changes position before and after the interaction: hand/arm, foot/leg, torso, head/face, and multiple parts (see Fig.~\ref{fig:dataset_sta}(c)). Hand/arm interactions are the most dominant (54.3\%), other interactions involve foot/leg (15.0\%), multiple parts (12.5\%), torso (11.0\%), and head/face (7.2\%). This distribution highlights the diversity of interaction types and the involved body regions in our dataset.

\textbf{(4) Foreground Object Size:} Our dataset includes foreground objects of varying sizes. Based on the ratio of foreground object area to the entire image area, we classify them into three categories: small (<10\%), medium (10--30\%), and large (>30\%) (see Fig.~\ref{fig:dataset_sta}(d)). Medium objects are the most common (44.3\%), followed by small (29.2\%) and large (26.5\%). This distribution indicates that our dataset captures a diverse range of object sizes, which is essential for evaluating interaction robustness across different foreground scales.

\textbf{(5) Image Style:} Our dataset spans five distinct image styles: photo-realistic, cartoon, sketch-like, artistic, and digital art (see Fig.~\ref{fig:dataset_sta}(e)). Photo-realistic images comprise the majority (65.8\%), while the remaining styles each account for 8.5\%. This diversity supports our method in handling images from different visual domains.

\textbf{(6) Background Scene Type:} Our dataset includes images with diverse background scenes, which we use GPT-4o to judge the complexity of background scene: simple indoor, complex indoor, simple outdoor, and complex outdoor (see Fig.~\ref{fig:dataset_sta}(f)). The distribution is relatively balanced: complex indoor (29.2\%), simple indoor (27.9\%), complex outdoor (23.3\%), and simple outdoor (19.6\%), ensuring broad coverage across varied scene contexts.
 
\section{Effectiveness of Residual-based Modulation Strategy}
\label{supp:attention}
As discussed in Sec. 3.3 of the main paper, our shape-aware attention modulation employs a residual-based strategy to adjust the attention maps. This design is motivated by the concern that directly modifying attention maps may degrade the visual quality of the generated images, as suggested by previous work~\cite{kim2023dense}.

\begin{figure}[h]
\centering
\includegraphics[width=1\linewidth]{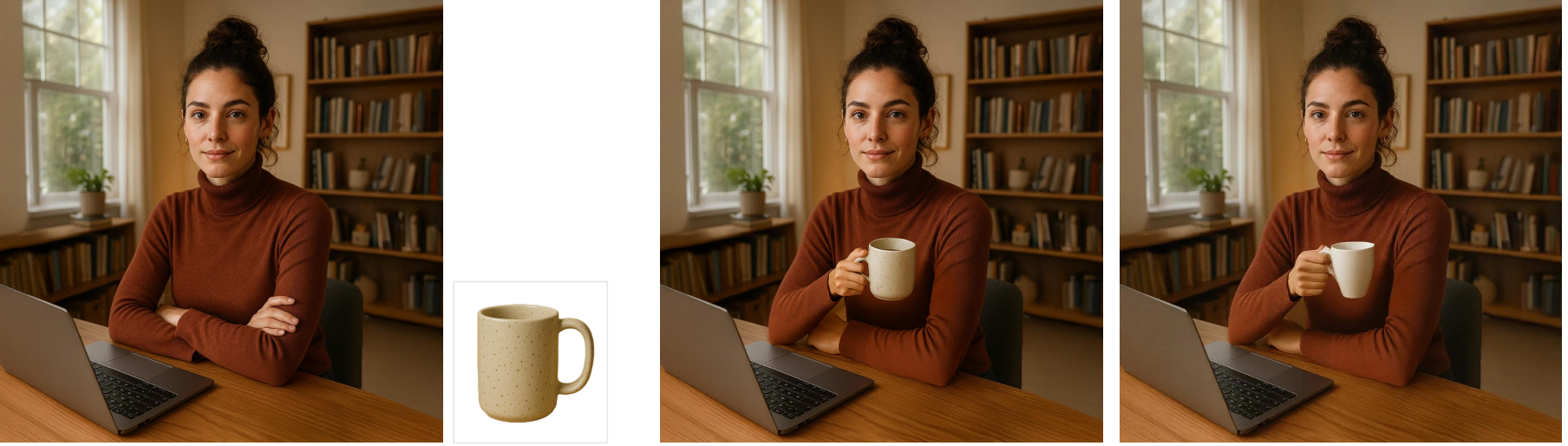}
  \put(-360,120){\small{(a)Input image \& object}}
  \put(-205,120){\small{(b)Residual Strategy}}
  \put(-100,120){\small{(c)Non-residual Strategy}}
\caption{Visual results of ablation study on attention modulation strategies in Table~\ref{tab:attention_str}.}
\label{fig:attention_str}
\end{figure}

We define our modulation as:
\[
A' = A + \alpha \cdot \left( M_{\text{shape}} \cdot (A_{\max} - A) - (1 - M_{\text{shape}}) \cdot (A - A_{\min}) \right)
\]
where $A$ is the original attention map, $M_{\text{shape}}$ is the ground-truth shape mask, $\alpha$ is a modulation strength, $A_{\max}$ and $A_{\min}$ denote the maximum and minimum attention values per query. The terms $(A_{\max} - A)$ and $(A - A_{\min})$ serve as residuals, which helps constrain the modulation within the original attention range. This ensures that the updated attention map $A'$ does not deviate excessively, thereby preserving the pretrained model’s attention distribution.
For comparison, we also evaluate a naive modulation strategy without residual constraints, formulated as:
\[
A' = A + \alpha \cdot \left( M_{\text{shape}} - (1 - M_{\text{shape}}) \right)
\]
We conduct an ablation study on the HOIBench to compare the effectiveness of the residual-based strategy versus the non-residual version. As shown in \figref{fig:attention_str} and Table.~\ref{tab:attention_str}, removing the residual leads to a notable drop in FID and DINO scores, indicating degraded image quality and reduced consistency of the generated foreground objects. Other metrics also show minor decreases. Visually, the generated shapes deviate more from the input guidance, confirming the importance of the residual design.

\begin{table}[h]
\centering
\small
\caption{Ablation study on attention modulation strategies.}
\vspace{0mm}
\renewcommand\arraystretch{1.1}
\scalebox{0.85}{
\begin{tabular}{lccccc}
\toprule
\textbf{Modulation Strategy} & \textbf{FID} $\downarrow$ & \textbf{CLIP} $\uparrow$ & \textbf{HOI} $\uparrow$ & \textbf{DINO} $\uparrow$ & \textbf{SSIM(BG)} $\uparrow$ \\
\midrule
Non-residual Strategy & 10.89 & 30.07 & 84.32 & 69.72 & 95.58 \\
Residual Strategy     & \textbf{9.27}  & \textbf{30.29} & \textbf{87.39} & \textbf{78.21} & \textbf{96.57} \\
\bottomrule
\end{tabular}
}
\label{tab:attention_str}
\end{table}

\vspace{0mm}

\section{Effect of Coefficients}

We evaluate the impact of four coefficients in the overall training loss and the shape-aware attention modulation on HOIBench. Specifically,
$\alpha_1$, $\alpha_2$, and $\alpha_3$ are the coefficients of the pose-guided loss, background consistency loss, and multi-view appearance loss, respectively. $\alpha$ denotes the modulation strength used in the shape-aware attention modulation.

As shown in Table.~\ref{tab:loss_comparison}. \ding{182}Increasing $\alpha_1$ from 1 to 1.5 (Rows 1 vs. 2) improves HOI score (87.39 $\rightarrow$ 88.01) and CLIP score (30.29 $\rightarrow$ 30.31), indicating better pose alignment. However, this comes at the cost of image quality and consistency, with FID increasing (9.27 $\rightarrow$ 10.65), and both DINO and SSIM(BG) decreasing (78.21 $\rightarrow$ 73.32, 96.57 $\rightarrow$ 94.33). \ding{183}Raising $\alpha_2$ from 0.5 to 1.0 (Rows 1 vs. 3) improves SSIM(BG) (96.57 $\rightarrow$ 96.92), reflecting better background preservation, but significantly degrades other metrics including FID, CLIP, HOI, and DINO—suggesting that excessive emphasis on background stability impairs semantic and visual coherence. \ding{184}Increasing $\alpha_3$ from 0.8 to 1.0 (Rows 1 vs. 4) slightly improves DINO (78.21 $\rightarrow$ 78.58), indicating enhanced shape alignment, but at the cost of higher FID (12.92) and lower SSIM(BG) (94.88), showing a trade-off between appearance consistency and image quality. \ding{185}Finally, increasing modulation strength $\alpha$ from 1.0 to 1.5 (Rows 1 vs. 5) causes moderate declines in FID (9.27 $\rightarrow$ 10.87), DINO (78.21 $\rightarrow$ 77.63), and SSIM(BG) (96.57 $\rightarrow$ 95.48), this effect may arise due to the destabilization of the pretrained attention distribution caused by excessively aggressive attention modulation.

\begin{table}[h]
    \centering
    \caption{Quantitative comparison of different coefficient combinations. $\alpha_1$, $\alpha_2$, and $\alpha_3$ are the coefficients of the pose-guided loss, background consistency loss, and multi-view appearance loss, respectively. $\alpha$ denotes the modulation strength used in the shape-aware attention modulation.}
    \vspace{2mm}
    \label{tab:loss_comparison}
    \renewcommand\arraystretch{1.3}
    \scalebox{0.9}{
    \begin{tabular}{lccccc}
        \toprule
        \textbf{Coefficients ($\alpha_1$, $\alpha_2$, $\alpha_3$, $\alpha$)} & \textbf{FID~$\downarrow$} & \textbf{CLIP~$\uparrow$} & \textbf{HOI~$\uparrow$} & \textbf{DINO~$\uparrow$} & \textbf{SSIM(BG)~$\uparrow$} \\
        \midrule
        $\alpha_1{=}1$, $\alpha_2{=}0.5$, $\alpha_3{=}0.8$, $\alpha{=}1$   & \textbf{9.27}  & \textbf{30.29} & \textbf{87.39} & \textbf{78.21} & \textbf{96.57} \\
        $\alpha_1{=}1.5$, $\alpha_2{=}0.5$, $\alpha_3{=}0.8$, $\alpha{=}1$ & 10.65 & 30.31 & 88.01 & 73.32 & 94.33 \\
        $\alpha_1{=}1$, $\alpha_2{=}1$, $\alpha_3{=}0.8$, $\alpha{=}1$     & 11.29 & 29.88 & 82.16 & 74.10 & 96.92 \\
        $\alpha_1{=}1$, $\alpha_2{=}0.5$, $\alpha_3{=}1$, $\alpha{=}1$     & 12.92 & 29.71 & 85.75 & 78.58 & 94.88 \\
        $\alpha_1{=}1$, $\alpha_2{=}0.5$, $\alpha_3{=}0.8$, $\alpha{=}1.5$ & 10.87 & 30.25 & 86.11 & 77.63 & 95.48 \\
        \bottomrule
    \end{tabular}
    }
\end{table}

\section{Extended Details on Using MLLMs to Identify Interaction Types and Regions}

In Sec. 3.2 of the main paper, we briefly described the use of MLLMs to infer interaction types and interaction regions via multi-turn querying. Here, we detail the full process.

Given a background human image $I_b$ and a foreground object image $I_f$, we iteratively use an MLLM to extract: (1) a text prompt $C$ describing the interaction, (2) the object bounding box $B_o$, and (3) the interaction region on the human $B_r$. The multi-turn procedure proceeds as follows:

\begin{enumerate}[label=\arabic*., left=0pt]
    \item \textbf{Interaction Prompt Generation.}  
    The MLLM is queried with $I_f$ and $I_b$ using the instruction: \textit{``Please analyze and describe a suitable type of interaction between them and generate a simple prompt for this interaction.''} The model outputs a text prompt $C$ describing the interaction type.
    
    \item \textbf{Object Box Prediction.}  
    Using $I_f$, $I_b$, and $C$, we query the MLLM with: \textit{``Please describe the position of the foreground object and give bounding box coordinates so that it aligns with the specified interaction.''} The model returns the object bounding box $B_o$.
    
    \item \textbf{Interaction Region Prediction.}  
    Given $I_f$, $I_b$, $C$, and $B_o$, we ask: \textit{``Based on the images and interaction prompt, and assuming the object is at $B_o$, identify the regions on the person that would be affected during the interaction and return their bounding box.''} The MLLM then predicts the interaction region box $B_r$.
\end{enumerate}

\section{Additional Ablation studies}
\label{supp:ablation}
\subsection{Multi-View Generators and View Numbers}

\begin{figure}[h]
\centering
\includegraphics[width=1\linewidth]{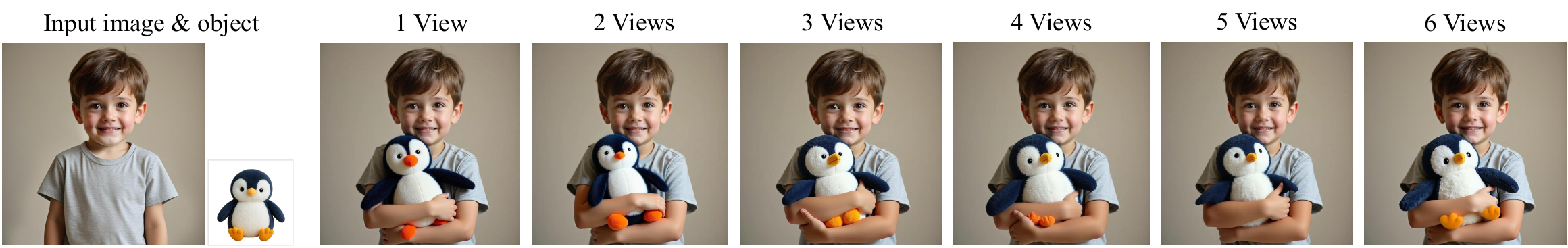}
\caption{Visual results of ablation study on view numbers used in multi-view appearance loss.}
\label{fig:abl_num_view}
\end{figure}

We evaluate the impact of the number of views used in the multi-view appearance loss (\figref{fig:abl_num_view}, Table.~\ref{tab:abl_multi_view} (left)). Using only a single view leads to noticeable inconsistencies in object appearance. As the number of views increases, performance improves steadily across all metrics, confirming the value of richer multi-view supervision.

\begin{figure}[h]
\centering
\includegraphics[width=1\linewidth]{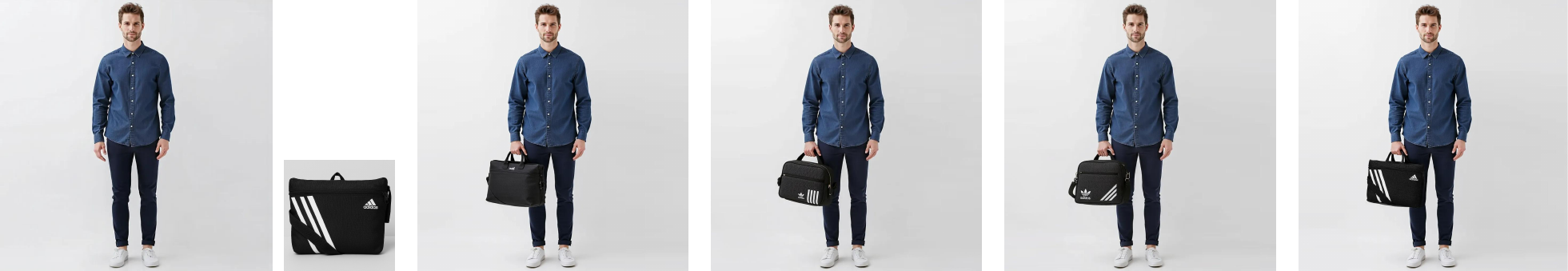}
\put(-375,75){\scriptsize{Input image \& object}}
\put(-277,75){\scriptsize{No Multi-view}}
\put(-195,75){\scriptsize{SV3D~\cite{voleti2024sv3d}}}
\put(-125,75){\scriptsize{ViewDiff~\cite{hollein2024viewdiff}}}
\put(-50,75){\scriptsize{Zero123+~\cite{shi2023zero123++}}}
\caption{Visual results of ablation study on multi-view generators.}
\label{fig:abl_multi_view}
\vspace{-5mm}
\end{figure}

We further evaluate different multi-view generation methods (\figref{fig:abl_multi_view}, Table.~\ref{tab:abl_multi_view} (right)). Without multi-view supervision, the model fails to maintain appearance consistency under significant viewpoint changes. Incorporating multiple generated views into the CLIP loss enhances coherence across varying poses and backgrounds. Among the methods, Zero123+\cite{parmar2023zero} achieves the best results, while SV3D\cite{voleti2024sv3d} and ViewDiff~\cite{hollein2024viewdiff} also outperform the no multi-view baseline, underscoring the importance of high-fidelity multi-view supervision.

\vspace{-3mm}

\vspace{-2mm}

\begin{table}[h]
\small
\centering
\caption{Ablation on different numbers of views (left) and multi-view generators (right).}
\renewcommand\arraystretch{1.1}
\begin{minipage}{0.47\linewidth}
\centering
\scalebox{0.7}{
\begin{tabular}{lccccc}
\toprule
\textbf{\# Views} & \textbf{FID} $\downarrow$ & \textbf{CLIP} $\uparrow$ & \textbf{HOI} $\uparrow$ & \textbf{DINO} $\uparrow$ & \textbf{SSIM(BG)} $\uparrow$ \\
\midrule
1(No multi-view) & 11.55 & 29.52 & 81.32 & 68.83 & 95.83 \\
2 & 10.22 & 29.55 & 83.89 & 69.73 & 95.86 \\
3 & 10.19 & 29.81 & 85.08 & 70.26 & 95.87 \\
4 & 9.54  & 30.21 & 85.19 & 71.63 & 96.03 \\
5 & 9.29  & 30.23 & 86.07 & 74.19 & 96.21 \\
\textbf{6} & \textbf{9.27}  & \textbf{30.29} & \textbf{87.39} & \textbf{78.21} & \textbf{96.57} \\
\bottomrule
\end{tabular}
}
\end{minipage}
\hfill
\renewcommand\arraystretch{1.53}
\begin{minipage}{0.47\linewidth}
\centering
\scalebox{0.7}{
\begin{tabular}{l|ccccc}
\toprule
\textbf{Method} & \textbf{FID} $\downarrow$ & \textbf{CLIP} $\uparrow$ & \textbf{HOI} $\uparrow$ & \textbf{DINO} $\uparrow$ & \textbf{SSIM(BG)} $\uparrow$ \\
\midrule
No multi-view & 11.55 & 29.52 & 81.32 & 68.83 & 95.83 \\
\textbf{Zero123+}\cite{parmar2023zero} & \textbf{9.27} & \textbf{30.29} & \textbf{87.39} & \textbf{78.21} & \textbf{96.57} \\
SV3D\cite{voleti2024sv3d} & 9.89 & 29.85 & 84.98 & 75.26 & 96.01 \\
ViewDiff\cite{hollein2024viewdiff} & 10.20 & 29.99 & 86.19 & 74.63 & 95.98 \\
\bottomrule
\end{tabular}
}
\end{minipage}
\label{tab:abl_multi_view}
\vspace{-3mm}
\end{table}

\subsection{LoRA Ranks}

\begin{wraptable}{r}{0.4\linewidth}
\vspace{1mm}
\centering
\small
\caption{Ablation study on LoRA Ranks}
\renewcommand\arraystretch{1.3}
\scalebox{0.65}{
\begin{tabular}{lccccc}
\toprule
\textbf{Rank} & \textbf{FID} $\downarrow$ & \textbf{CLIP} $\uparrow$ & \textbf{HOI} $\uparrow$ & \textbf{DINO} $\uparrow$ & \textbf{SSIM(BG)} $\uparrow$ \\
\midrule
8   & 9.51 & 29.98 & 84.32 & 74.72 & 96.12 \\
\textbf{16}  & \textbf{9.27} & \textbf{30.29} & \textbf{87.39} & \textbf{78.21} & \textbf{96.57} \\
32  & 9.84 & 30.24 & 86.68 & 77.26 & 96.15 \\
64  & 9.33 & 30.27 & 85.49 & 77.12 & 96.04 \\
\bottomrule
\end{tabular}
}
\label{tab:abl_lora_rank}
\vspace{5mm}
\end{wraptable}

Table~\ref{tab:abl_lora_rank} presents the results of varying the LoRA rank (8, 16, 32, 64) across five evaluation metrics.
Rank 16 consistently achieves the best overall performance, yielding the lowest FID (9.27) and the highest scores in CLIP (30.29), HOI (87.39), DINO (78.21), and SSIM(BG) (96.57). When the rank is too low (e.g., 8), the model underperforms across all metrics, indicating insufficient capacity to model human-object interactions and maintain consistent appearances. However, higher ranks (32, 64) yield marginal or no improvements (\eg, DINO drops to 77.26 and 77.12), suggesting possible overfitting.

\subsection{ID Encoder Backbone}

As discussed in Sec. 3.3 of the main paper, we adopt DINOv2 as the backbone for extracting object identity features. Here, we conduct an ablation study comparing different backbones: VAE~\cite{navard2024knobgen}, CLIP~\cite{safaee2024clic}, and DINOv2~\cite{liu2023grounding}. To ensure a fair evaluation, we additionally report CLIP-I~\cite{radford2021learning}, which measures the CLIP similarity between the synthesized foreground object and the input foreground object.

\begin{figure}[h]
\centering
\includegraphics[width=1\linewidth]{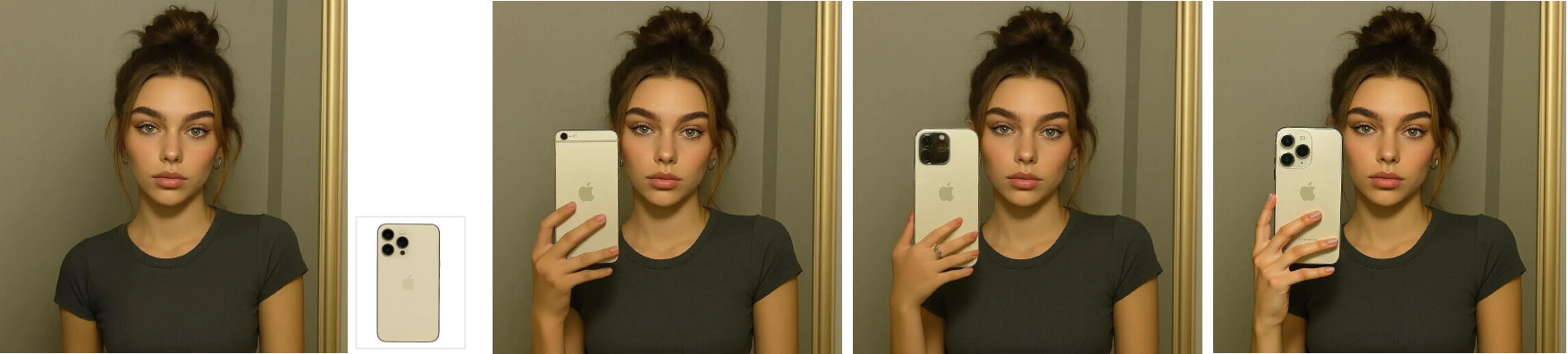}
\put(-370,95){\small{Input image \& object}}
\put(-240,95){\small{VAE\cite{kingma2013auto}}}
\put(-150,95){\small{CLIP\cite{safaee2024clic}}}
\put(-65,95){\small{DINOv2\cite{oquabdinov2}}}
\caption{Ablation study on different backbones for foreground ID encoders.}
\label{fig:ab_encoder}
\vspace{-3mm}
\end{figure}
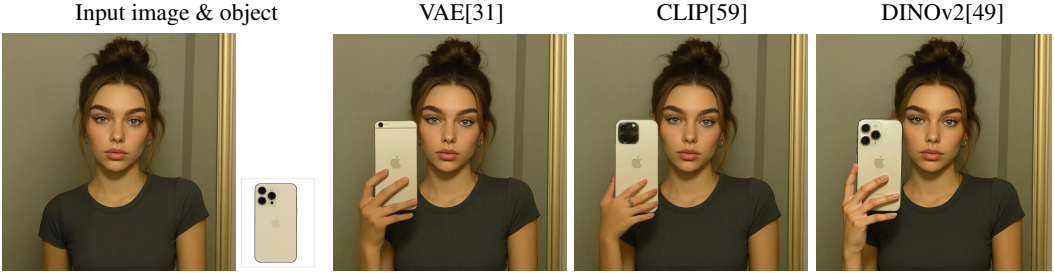

As shown in Table.~\ref{tab:ab_encoder_transpose}, DINOv2 consistently outperforms other ID encoder backbones across all evaluated metrics. As shown in Fig.~\ref{fig:ab_encoder}, using DINOv2 as the ID encoder backbone yields the most consistent foreground object.

\begin{table}[h]
\small
\caption{Ablation study on different ID encoder backbones}
\renewcommand\arraystretch{1.2}
\centering
\scalebox{0.7}{
\begin{tabular}{l|cccccc}
\toprule
\textbf{Backbone} & \textbf{FID} $\downarrow$ & \textbf{CLIP} $\uparrow$ & \textbf{HOI} $\uparrow$ & \textbf{DINO} $\uparrow$ & \textbf{CLIP-I} $\uparrow$ & \textbf{SSIM(BG)} $\uparrow$ \\
\midrule
VAE~\cite{navard2024knobgen}   & 9.98 & 29.72 & 82.73 & 67.33 & 78.38 & 95.98 \\
CLIP~\cite{safaee2024clic}     & 9.55 & 30.17 & 85.24 & 75.72 & 87.79 & 96.53 \\
\textbf{DINOv2}~\cite{liu2023grounding} & \textbf{9.27} & \textbf{30.29} & \textbf{87.39} & \textbf{78.21} & \textbf{90.25} & \textbf{96.57} \\
\bottomrule
\end{tabular}
}
\label{tab:ab_encoder_transpose}
\end{table}

\subsection{Guidance Scale}

To study the impact of the guidance scale on our model, we evaluate performance under six different inference-time guidance scales: 1, 2, 3, 3.5, 4, and 5.

\begin{wraptable}{r}{0.45\textwidth}
\centering
\renewcommand{\arraystretch}{1.1}
\scalebox{0.58}{
\begin{tabular}{lccccc}
\toprule
\textbf{Guidance Scale} & \textbf{FID~$\downarrow$} & \textbf{CLIP~$\uparrow$} & \textbf{HOI~$\uparrow$} & \textbf{DINO~$\uparrow$} & \textbf{SSIM(BG)~$\uparrow$} \\
\midrule
gs = 1.0 & 10.11 & 29.42 & 80.01 & 62.33 & 95.25 \\
gs = 2.0 & 9.78 & 29.85 & 81.56 & 71.60 & 95.28 \\
gs = 3.0 & 9.48 & 30.12 & 82.47 & 74.04 & 95.21 \\
\textbf{gs = 3.5} & \textbf{9.27} & \textbf{30.29} & \textbf{87.39} & \textbf{78.21} & \textbf{96.57} \\
gs = 4.0 & 9.39 & 30.19 & 83.91 & 77.56 & 95.89 \\
gs = 5.0 & 9.68 & 29.76 & 81.23 & 76.41 & 96.18 \\
\bottomrule
\end{tabular}
}
\caption{Performance of our model under different guidance scales during inference. The model is trained with a guidance scale of 1.}
\vspace{3mm}
\label{tab:guidance_scale}
\end{wraptable}

\begin{figure}[h]
\centering
\includegraphics[width=1\linewidth]{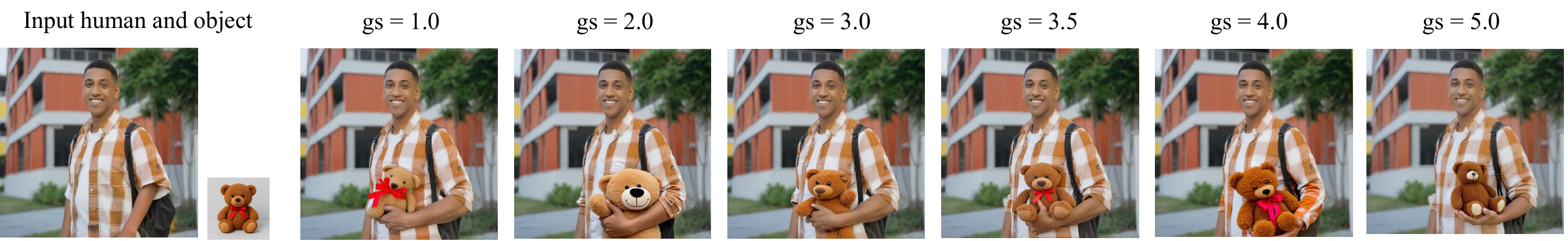}
\caption{Ablation study on different guidance scales (denoted as gs) during inference.}
\label{fig:abl_guidance_scale}
\vspace{-3mm}
\end{figure}

As shown in Table.~\ref{tab:guidance_scale} and Fig.~\ref{fig:abl_guidance_scale}, guidance scale = 3.5 achieves the best overall performance (FID = 9.27, CLIP = 30.29, HOI = 87.39, DINO = 78.21, SSIM(BG) = 96.57). Correspondingly, the visual results at this setting exhibit the most faithful preservation of the foreground object's appearance. In contrast, lower guidance scales (gs = 1.0 or 2.0) lead to diminished semantic alignment, particularly evident in the foreground regions, as reflected by lower DINO scores. Increasing the scale beyond 3.5 (\eg, gs = 4.0 or 5.0) results in subtle declines in both quantitative scores and foreground object consistency. 

\section{Comparison with Multi-Modality Models}
\label{supp:compare_MLLM}
We compare our method with recent state-of-the-art multi-modality models, including \textit{GPT-4o}\cite{chatgpt}, \textit{Grok3}\cite{xai2025grok3}, and \textit{MidJourney V7}~\cite{midjourney2025}. All models receive identical inputs: a foreground object, a background human image, a designated interaction region, and a corresponding text prompt.

\begin{figure}[h]
\centering
\includegraphics[width=0.95\linewidth]{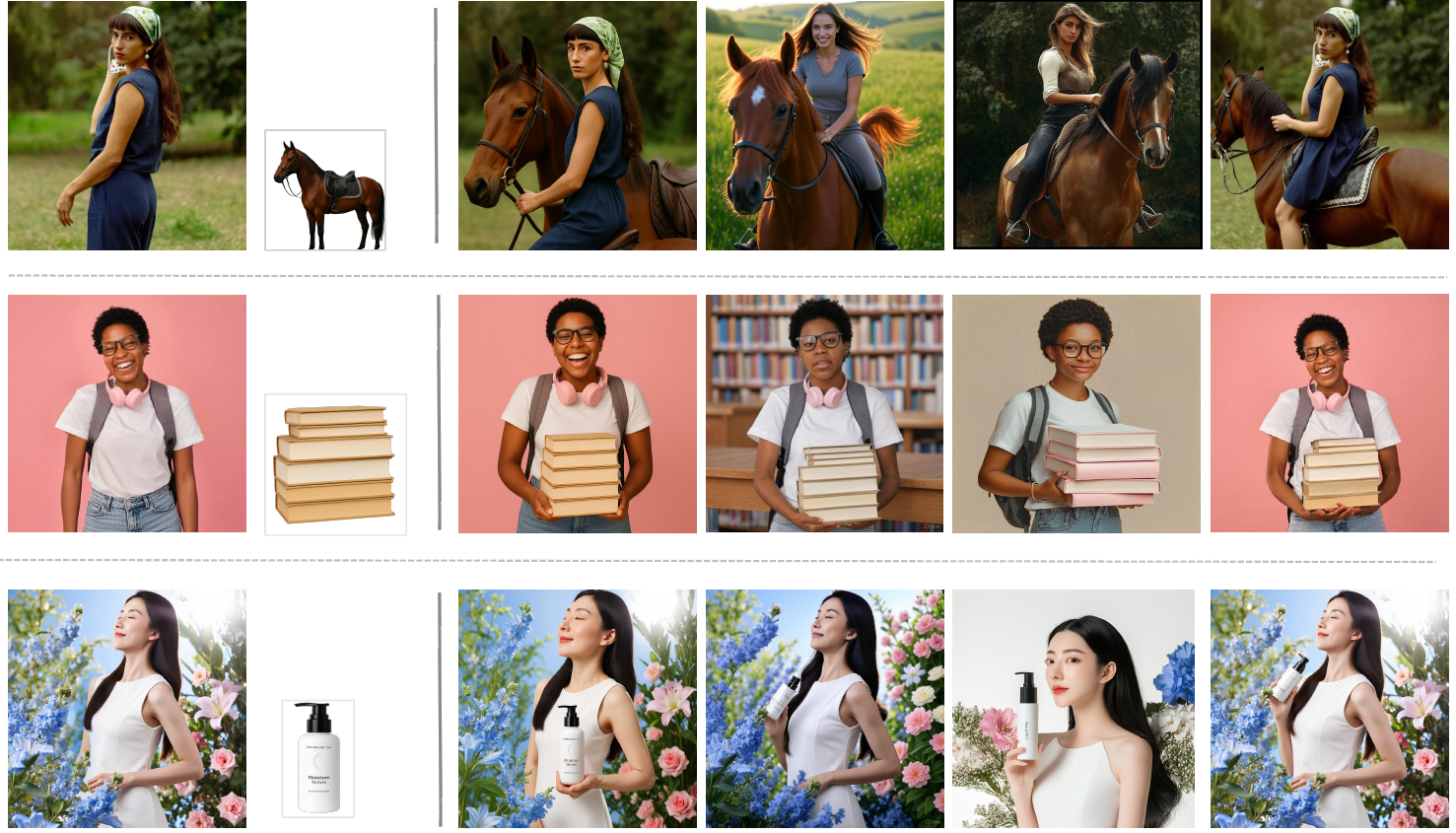}
\put(-365,225){\small{(a)Input image \& object}}
\put(-255,225){\small{(b)GPT-4o\cite{chatgpt}}}
\put(-190,225){\small{(c)Grok3\cite{xai2025grok3}}}
\put(-135,225){\small{(d)MidJourney V7\cite{midjourney2025}}}
\put(-45,225){\small{(e)Ours}}
\caption{Quantitative comparison with recent state-of-the-art multi-modality models. The prompts for the above three cases are: "A woman is riding a horse","A girl is holding a stack of books", "A model is presenting a skincare bottle".}
\label{fig:compare_MLLM}
\vspace{-3mm}
\end{figure}
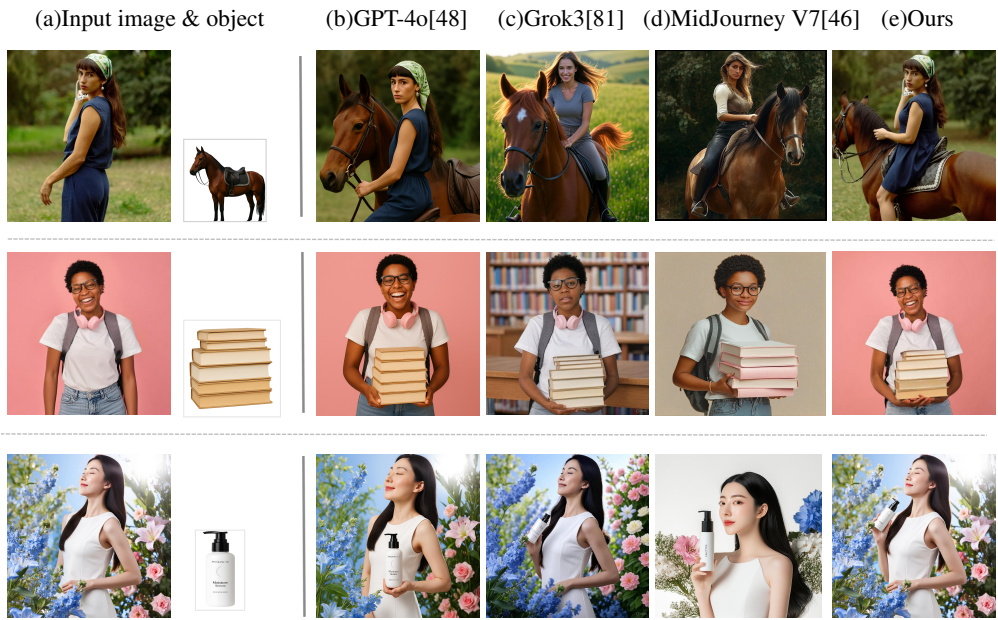

Qualitative results reveal clear limitations in existing models. GPT-4o and MidJourney V7 frequently fail to generate consistent foreground objects (\eg, Row 2(b), Rows 2–3(d) in Fig.~\ref{fig:compare_MLLM}). Grok3 and MidJourney V7 struggle to preserve the background human and scene details (Rows 1–3(c–d)). In addition, GPT-4o may struggle to accurately model interactions under complex scenarios (see Row 1(b)).

Quantitatively, our method outperforms all baselines across five key metrics. It achieves the lowest FID (9.27), highest CLIP score (30.29), HOI score (87.39), DINO score (78.21) and SSIM(BG) score (96.57). This demonstrate that our method delivers more harmonious human-object interactions and consistent appearances.

\begin{table}[h]
\centering
\caption{Qualitative comparison with recent state-of-the-art multi-modality models.}
\vspace{2mm}
\label{tab:gpt_comparison}
\renewcommand{\arraystretch}{1.2}
\scalebox{1}{
\begin{tabular}{lccccc}
\toprule
\textbf{Method} & \textbf{FID$\downarrow$} & \textbf{CLIP$\uparrow$} & \textbf{HOI$\uparrow$} & \textbf{DINO$\uparrow$} & \textbf{SSIM(BG)$\uparrow$} \\
\midrule 
Grok3~\cite{xai2025grok3} & 13.27 & 29.07 & 65.03 & 57.02 & 58.25 \\
GPT-4o~\cite{chatgpt} & 9.98 & 29.35 & 75.22 & 65.23 & 47.22 \\
MidJourney V7~\cite{midjourney2025} & 10.85 & 29.87 & 73.45 & 60.18 & 41.34 \\
\textbf{Ours} & \textbf{9.27} & \textbf{30.29} & \textbf{87.39} & \textbf{78.21} & \textbf{96.57} \\
\bottomrule
\end{tabular}
}
\end{table}

\section{Additional Comparison with Image Composition Methods}

In addition to the nine methods compared in the main paper, we conducted further comparisons with five additional state-of-the-art image composition methods: DreamFuse~\cite{huang2025dreamfuse}, InsertAnything~\cite{song2025insert}, MimicBrush~\cite{chen2024zero}, Bifrost~\cite{li2024textit} and DreamRelation~\cite{shi2025dreamrelation}. For fairness, all methods with publicly available training code were retrained or fine-tuned on our dataset.

Fig.~\ref{fig:more_results} shows qualitative comparisons.
DreamFuse and InsertAnything generate visually faithful foreground objects, but often fail to model realistic human-object interactions (see Rows 2–4 in Fig.\ref{fig:more_results}(b–c)).
DreamRelation produces interaction-like gestures, yet struggles to preserve the visual consistency of the foreground object and background human (Rows 1–4 in Fig.\ref{fig:more_results}(f)).
MimicBrush and Bifrost, on the other hand, produce neither convincing interactions nor accurate object appearances (Fig.~\ref{fig:more_results}(d–e)).
In contrast, our method generates diverse and harmonious interactions while maintaining the consistent appearance of both the foreground and the background.

Table.~\ref{tab:sota} provides quantitative results.
Our method achieves the best FID (9.27), CLIP-Score (30.29), HOI-Score (87.39), and DINO-Score (78.21), indicating superior image quality, semantic alignment, interaction quality and appearance consistency. User study results further validate our approach, ranking it highest in image quality (IQ), interaction harmonization (IH), and appearance preservation (AP), with all scores significantly outperforming other methods.

\begin{figure}[t]
\centering
\includegraphics[width=1\linewidth]{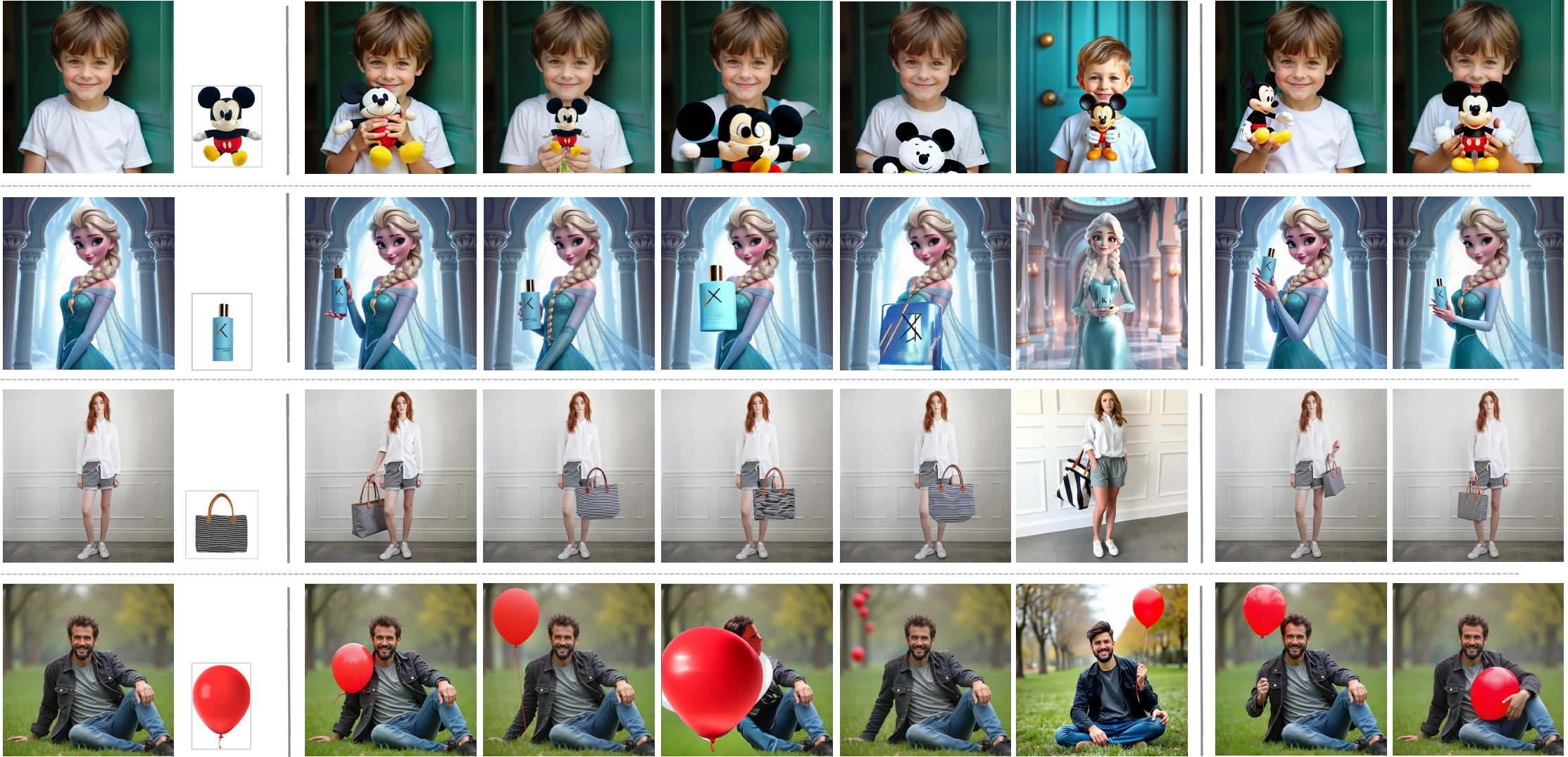}
\put(-400,195){\scriptsize{(a)Input image \& object}}
\put(-325,195){\scriptsize{(b)DreamFu.\cite{huang2025dreamfuse}}}
\put(-275,195){\scriptsize{(c)InsertAn.\cite{song2025insert}}}
\put(-226,195){\scriptsize{(d)Mimic.\cite{chen2024zero}}}
\put(-183,195){\scriptsize{(e)Bifrost\cite{li2024textit}}}
\put(-140,195){\scriptsize{(f)DreamRe.\cite{shi2025dreamrelation}}}
\put(-60,195){\scriptsize{(g)Ours}}
\caption{Additional qualitative comparisons of our \name~with 5 SOTA methods. The prompts for the above four examples are: “A boy is holding a mickey mouse toy”, “A girl is showing a perfume bottle”, “A woman is lifting a bag”, and “A sitting man is holding a balloon”.}
\label{fig:more_results}
\vspace{-3mm}
\end{figure}
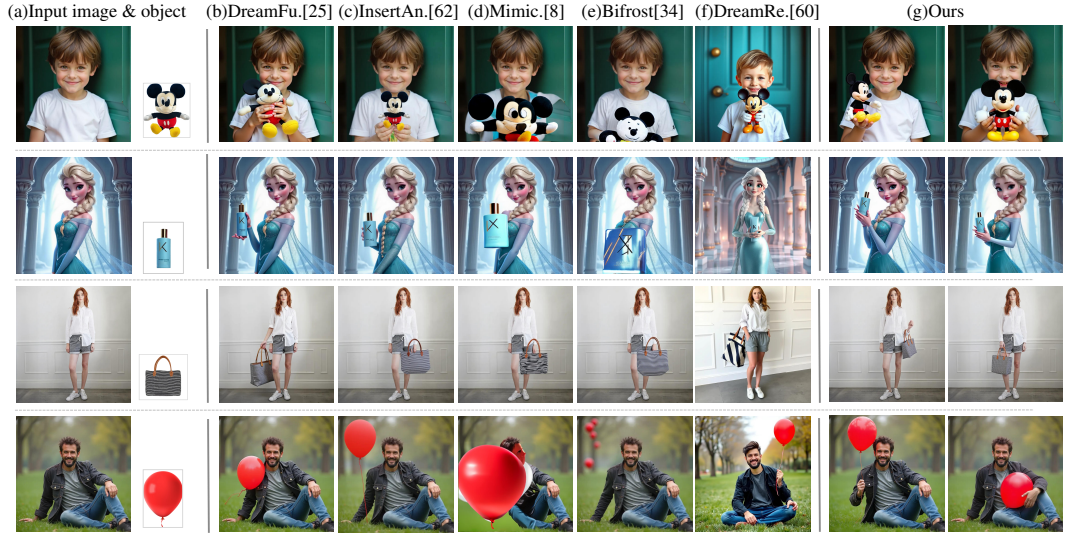

\begin{table*}[t]
\centering
\small
\caption{Additional quantitative comparison of our method with 5 SOTA methods. The best and second-best results are highlighted in \textbf{bold} and \underline{underline}, respectively. Training or tuning-based methods without released training codes are marked with a $^{\dag}$.}
\setlength{\tabcolsep}{3pt}
\renewcommand{\arraystretch}{1.1}
\scalebox{0.8}{
\begin{tabular}{l|l|cccccc}
\toprule
\multirow{1}{*}{\textbf{Category}} & \multirow{1}{*}{\textbf{Metrics}}
& DreamFuse$^{\dag}$~\cite{huang2025dreamfuse} & InsertAnything$^{\dag}$~\cite{song2025insert} & MimicBrush$^{\dag}$~\cite{chen2024zero}
& Bifrost$^{\dag}$~\cite{li2024textit} & DreamRelation~\cite{shi2025dreamrelation} & Ours \\
\midrule
\multirow{5}{*}{Automatic}
 & FID $\downarrow$          & 13.35 & \underline{10.72} & 15.88 & 16.21 & 15.85 & \textbf{9.27} \\
 & CLIP-Score $\uparrow$     & 29.53 & \underline{29.76} & 28.62 & 28.17 & 28.55 & \textbf{30.29} \\
 & HOI-Score $\uparrow$      & \underline{63.75} & 58.85 & 36.04 & 38.98 & 52.66 & \textbf{87.39} \\
 & DINO-Score $\uparrow$     & 44.89 & \underline{64.52} & 40.67 & 42.02 & 37.07 & \textbf{78.21} \\
 & SSIM(BG) $\uparrow$       & \underline{93.23} & 92.19 & 84.56 & 88.11 & 25.19 & \textbf{96.57} \\
\midrule
\multirow{3}{*}{User study}
 & IQ $\downarrow$           & 3.10 & \underline{2.88} & 4.80 & 5.25 & 3.85 & \textbf{1.12} \\
 & IH $\downarrow$           & \underline{2.28} & 2.43 & 6.00 & 5.95 & 3.27 & \textbf{1.07} \\
 & AP $\downarrow$           & 2.89 & \underline{2.43} & 4.33 & 4.44 & 5.90 & \textbf{1.01} \\
\bottomrule
\end{tabular}}
\label{tab:sota}
\end{table*}

\section{Additional Results of \name}
\label{supp:more_results}
Fig.~\ref{fig:more_results} shows additional qualitative results of our method.
Each example includes: (1) Top: the final composited image, (2) Bottom: the input background human and foreground object.
These results demonstrate that our method produces natural and plausible human-object interactions while maintaining visual consistency of both the foreground object and the background human.

\begin{figure}[h]
\centering
\includegraphics[width=1\linewidth]{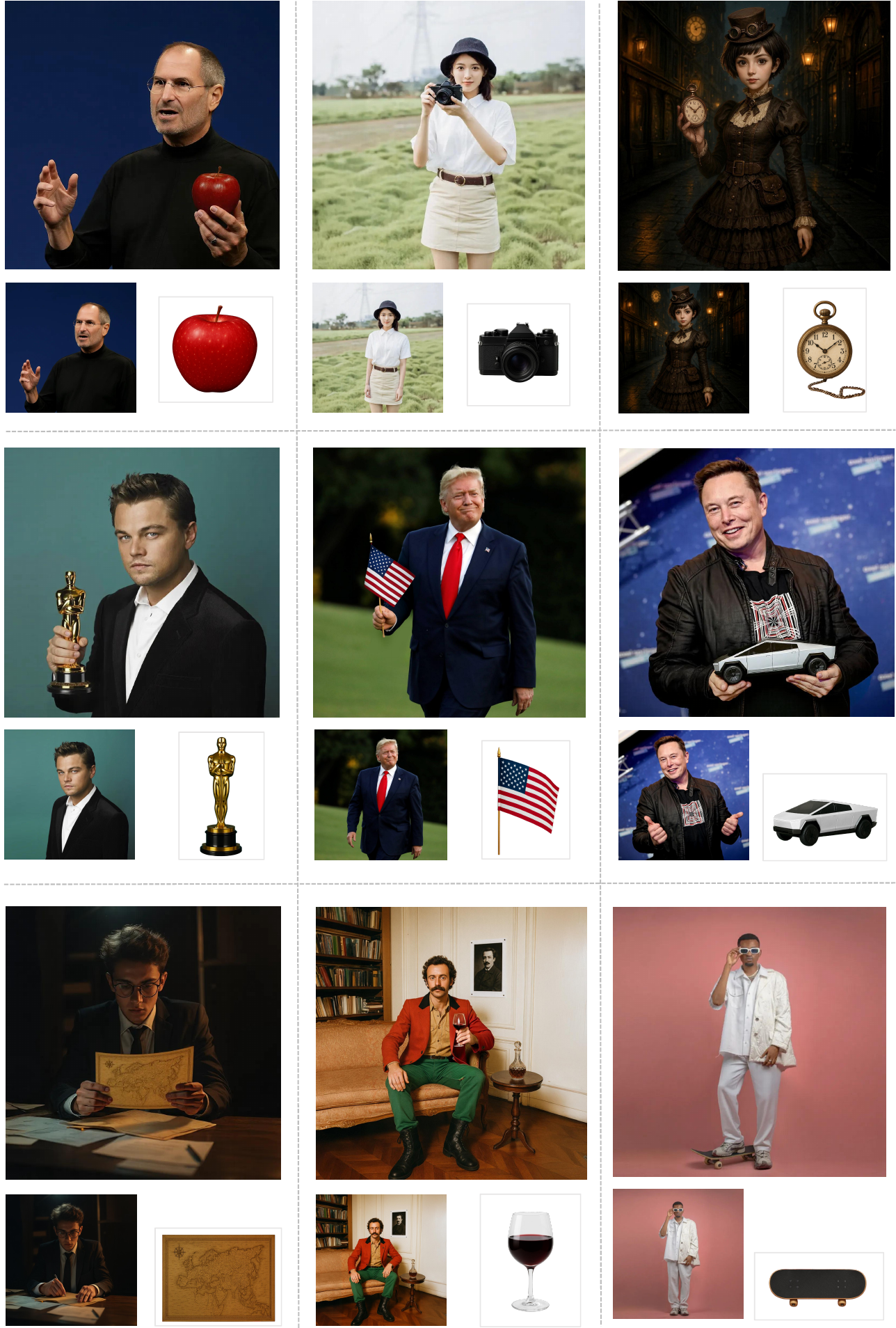}
\vspace{-3mm}
\caption{Additional qualitative results of \name. Each example includes: (1) Top: the final composited image, (2) Bottom: the input background human and foreground object. }
\label{fig:more_results}
\vspace{-3mm}
\end{figure}

\end{document}